\documentclass[10pt,twocolumn,letterpaper]{article}

\usepackage[pagenumbers]{cvpr} 

\usepackage{blindtext}
\usepackage{duckuments}
\usepackage{multirow}

\usepackage{pifont}
\definecolor{Gray}{gray}{0.9} 

\usepackage{pifont}
\usepackage{xcolor}
\usepackage{makecell}
\usepackage{subcaption}
\usepackage[table]{xcolor}
\usepackage{colortbl}
\usepackage{multirow, makecell}
\usepackage{wrapfig}
\newcommand{\paragrapht}[1]{\vspace{-10pt}\paragraph{#1}}
\definecolor{shcolor}{rgb}{0,0,1}

\usepackage{algorithm}
\usepackage{listings}
\usepackage{xcolor}
\usepackage{algpseudocode}
\definecolor{vsblue}{RGB}{0, 51, 179}       
\definecolor{vsgreen}{RGB}{0, 128, 0}       
\definecolor{vspurple}{RGB}{121, 94, 38}    
\definecolor{vsstring}{RGB}{163, 21, 21}    
\definecolor{vsnumber}{RGB}{9, 134, 88}     
\definecolor{vsgray}{RGB}{128, 128, 128}    

\lstdefinestyle{vscode}{
    language=Python,
    basicstyle=\ttfamily\footnotesize,      
    columns=fullflexible,                   
    keepspaces=true,                        
    keywordstyle=\color{vsblue}\bfseries,   
    commentstyle=\color{vsgreen}\itshape,   
    stringstyle=\color{vsstring},           
    numbers=left,                           
    numberstyle=\tiny\color{vsgray},        
    numbersep=5pt,                          
    tabsize=4,
    breaklines=true,
    showstringspaces=false,
    frame=none,                             
    emph={VGGT, VGGT_F, normalize, einsum, filter},
    emphstyle=\color{vspurple}\bfseries
}
    \newcommand\blfootnote[1]{
        \begingroup
        \renewcommand\thefootnote{}\footnote{#1}
        \addtocounter{footnote}{-1}
        \endgroup
    }

\lstdefinestyle{pytorch}{
  language=Python,
  basicstyle=\ttfamily\footnotesize, 
  keywordstyle=\ttfamily\footnotesize, 
  commentstyle=\itshape\color{gray},   
  stringstyle=\color{black},
  showstringspaces=false,
  tabsize=4,
  breaklines=true,
  emph={class,def,return},
  emphstyle=\bfseries\color{red},
}
\definecolor{cvprblue}{rgb}{0.21,0.49,0.74}
\usepackage[pagebackref,breaklinks,colorlinks,allcolors=cvprblue]{hyperref}

\def\confName{CVPR}
\def\confYear{2026}

\title{Emergent Outlier View Rejection in Visual Geometry Grounded Transformers}

\author{
    Jisang Han\textsuperscript{\rm 1,2}$^{\ddagger *}$ \quad
    Sunghwan Hong\textsuperscript{\rm 3}$^*$ \quad
    Jaewoo Jung\textsuperscript{\rm 1} \quad
    Wooseok Jang\textsuperscript{\rm 1} \quad
    Honggyu An\textsuperscript{\rm 1} \\ 
    Qianqian Wang\textsuperscript{\rm 4} \quad
    Seungryong Kim\textsuperscript{\rm 1}$^{\dagger}$ \quad
    Chen Feng\textsuperscript{\rm 2}$^{\dagger}$ \\[10pt]
    \textsuperscript{\rm 1}KAIST AI \qquad \textsuperscript{\rm 2}New York University \qquad \textsuperscript{\rm 3}ETH AI Center, ETH Zurich \qquad \textsuperscript{\rm 4}UC Berkeley\\
{\tt \href{https://cvlab-kaist.github.io/RobustVGGT}{\textcolor{purple}{https://cvlab-kaist.github.io/RobustVGGT}}}
}

\begin{document}
\maketitle
\blfootnote{$^\ddagger$Work done during a visiting research stay at New York University.}
\blfootnote{$^*$Equal contributions.}
\blfootnote{$^\dagger$Co-corresponding and equal advising.}


\begin{abstract}
Reliable 3D reconstruction from in-the-wild image collections is often hindered by ``noisy'' images—irrelevant inputs with little or no view overlap with others. While traditional Structure-from-Motion pipelines handle such cases through geometric verification and outlier rejection, feed-forward 3D reconstruction models lack these explicit mechanisms, leading to degraded performance under in-the-wild conditions. In this paper, we discover that the existing feed-forward reconstruction model, e.g., VGGT, despite lacking explicit outlier-rejection mechanisms or noise-aware training, can inherently distinguish distractor images. Through an in-depth analysis under varying proportions of synthetic distractors, we identify a specific layer that naturally exhibits outlier-suppressing behavior. Further probing reveals that this layer encodes discriminative internal representations that enable an effective noise-filtering capability, which we simply leverage to perform outlier-view rejection in feed-forward 3D reconstruction without any additional fine-tuning or supervision. Extensive experiments on both controlled and in-the-wild datasets demonstrate that this implicit filtering mechanism is consistent and generalizes well across diverse scenarios. 

\end{abstract}    
\section{Introduction}
\label{sec:intro}

\begin{figure}[t]
    \centering
    \includegraphics[width=1\linewidth]{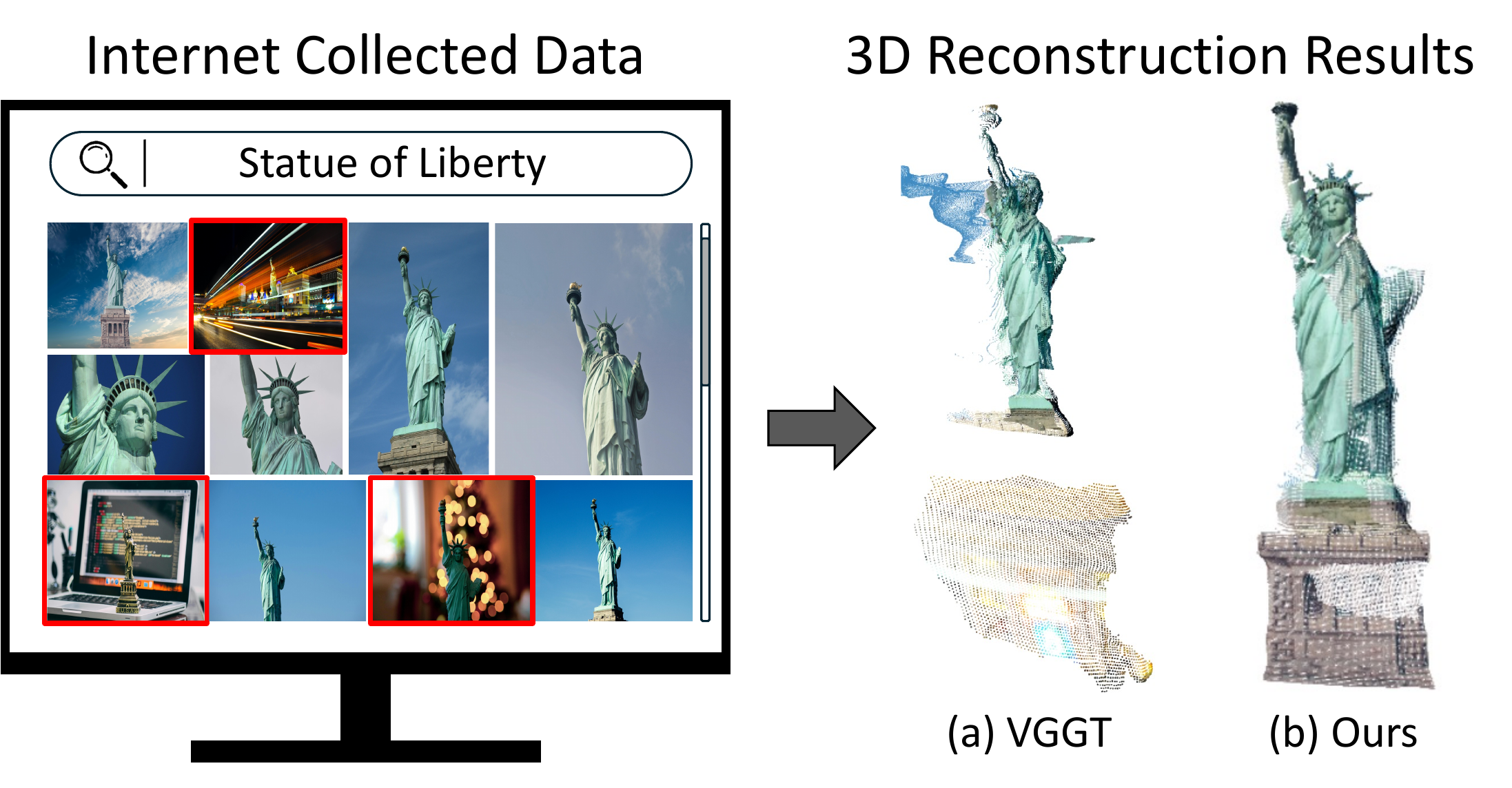}
    \caption{\textbf{Motivation.} In practice, image sets gathered for 3D reconstruction, \textit{e.g.,} via keyword search, often contain distractors or entirely irrelevant photos. As illustrated in (a), leaving these images unfiltered contaminates the VGGT~\cite{wang2025vggt} pipeline, producing noisy geometry and visible artifacts in the final reconstruction. In contrast, our \textbf{\textit{training-free}} approach, dubbed RobustVGGT, filters views using internal representations within VGGT~\cite{wang2025vggt}, yielding cleaner, more stable reconstructions, as shown in (b). }\vspace{-10pt}
    \label{fig:motivation}
\end{figure}

Multi-view 3D reconstruction~\cite{hartley2003multiple,schonberger2016structure}, the task of recovering scene geometry and camera poses from images with overlapping views, has been a central problem in computer vision, enabling applications across autonomous systems~\cite{zhu2024llava}, scene understanding~\cite{yoon2025visual,cho2024cat,kim2025seg4diff,shin2024towards} and AR/VR~\cite{lee20253d}. As these applications scale, they demand reconstructions that are accurate, efficient, and robust across diverse capture conditions.

Recent feed-forward reconstruction models~\cite{wang2024dust3r,leroy2024grounding,wang2024vggsfm} have made significant progress in this task by predicting geometry directly from image sets using learned representations and attention mechanisms. In particular, models such as VGGT~\cite{wang2025vggt} and Pi3~\cite{wang2025pi} bypass many iterative components of classical pipelines~\cite{rublee2011orb,fischler1981random,triggs1999bundle}, achieving impressive speed and strong performance on curated benchmarks.

A missing piece, however, is robustness to image collections with \textit{noisy or irrelevant views}. In practical settings, image collections often contain distractors, off-topic frames,  occlusions, or transient objects. This arises in a variety of scenarios, from casual captures and community photo collections to keyword-based queries \textit{e.g.,} “Statue of Liberty”, as exemplified in Fig.~\ref{fig:motivation}. These models~\cite{wang2024dust3r,wang2025vggt} lack an explicit mechanism to identify and remove inconsistent views; consequently, distractors can pass through the feed-forward pipeline, biasing pose estimation and degrading the recovered 3D structure (see Fig.~\ref{fig:motivation}(a)).

Classical Structure-from-Motion (SfM) pipelines address this challenge through stage-wise filtering and verification. After image retrieval~\cite{arandjelovic2016netvlad}, local features are detected and matched~\cite{detone2018superpoint, lowe2004distinctive, dusmanu2019d2}, and mismatches are pruned via robust estimation and geometric checks ~\cite{fischler1981random, hartley2003multiple}, followed by bundle adjustment ~\cite{triggs1999bundle}. Systems like COLMAP~\cite{schonberger2016structure} include these multi-stage procedures and are resilient to outlier views. However, they rely on iterative optimization and modular stages, which can limit scalability and tight integration with learning-based pipelines.

Our key insight is that VGGT~\cite{wang2025vggt}, despite lacking explicit outlier rejection mechanism or noise-aware training, can be \textit{repurposed for robust view selection} by leveraging an emergent properties of their internal representations: attention maps and intermediate features naturally emphasize spatially relevant, 3D-consistent views while downweighting distractors, \textit{without} fine-tuning, retraining or modifying the model. Through controlled analysis, we observe that certain layer in VGGT consistently suppress views that are inconsistent with the scene geometry. 

Building on this observation, we introduce a simple scoring scheme, called \textbf{RobustVGGT}, that quantifies per-view relevance using these internal signals and filters distractors with a single fixed threshold that generalizes across diverse datasets. This procedure introduces \textbf{\textit{no additional parameters or supervision}} and improves robustness while \textbf{\textit{preserving}} the efficiency of feed-forward reconstruction.
We evaluate our framework on real-world datasets ~\cite{snavely2006photo, ren2024nerf, sabour2023robustnerf, schoeps2017cvpr}. Despite its simplicity, our method consistently outperforms all compared baselines across datasets and different noise levels. We also provide ablations and analyses to validate our design choices. 

Our contributions include: 
\begin{itemize}
    \item  For the first time, we reveal an emergent noise-suppressing behavior in VGGT’s internal attention and feature representations through layer-wise quantitative and qualitative analyses.
    \item We propose a simple, training-free filtering mechanism that requires no architectural changes and selects geometrically consistent views by thresholding internal attention/feature signals with a single global parameter shared across datasets.
    \item We demonstrate consistent gains over strong baselines across diverse benchmarks and noise settings.
\end{itemize}

\section{Related work}
\label{sec:relwork}

\paragraph{\textit{Tabula rasa} 3D reconstruction.}
Structure from Motion (SfM) is a long-standing problem in computer vision that aims to estimate camera parameters and reconstruct 3D structure from multiple images of a static scene captured from different viewpoints. Traditional SfM pipelines~\cite{schonberger2016structure, agarwal2011building,hartley2003multiple,snavely2008modeling} comprise multiple stages-keypoint detection and description~\cite{lowe2004distinctive,tola2008fast,rublee2011orb,detone2018superpoint,dusmanu2019d2}, matching~\cite{sarlin2020superglue,sun2021loftr,hong2024unifying2,hong2021deep,edstedt2024roma,hong2022neural,cho2021cats,cho2022cats++}, triangulation~\cite{hartley1997triangulation}, outlier filtering~\cite{brachmann2017dsac,barath2019magsac,wei2023generalized,barath2020magsac++} and bundle adjustment~\cite{triggs1999bundle}—that jointly refine both structure and camera poses through iterative optimization. A key advantage of these pipelines is their robustness to noisy or irrelevant images or points: stages such as geometric verification, epipolar consistency checks, and RANSAC-based outlier rejection~\cite{fischler1981random,barath2019magsac} effectively filter mismatched features and erroneous correspondences before final reconstruction. Frameworks like COLMAP have become standard for reliable large-scale 3D reconstruction under such mechanisms. In contrast, recent feed-forward 3D reconstruction models (e.g., VGGT~\cite{wang2025vggt}) omit explicit filtering and can degrade on web-sourced, in-the-wild image sets with distractors. However, we show these models still exhibit \emph{implicit} outlier-suppression signals; by probing such layers and leveraging them at inference to select/reject views, we improve robustness without reinstating a full SfM pipeline.\vspace{-5pt}

\paragrapht{Learning-based  3D reconstruction.}
Learning-based multi-view 3D reconstruction has largely advanced through cost volumes that infer per-view depth followed by fusion. Supervised methods~\cite{yao2018mvsnet,yao2019recurrent,gu2020cascade,wang2021patchmatchnet} achieve strong accuracy on curated benchmarks but typically assume clean inputs and rely on view selection heuristics rather than explicit outlier rejection. To reduce annotation needs, weak/self-supervised approaches embed geometric consistency into the training or inference loop, \textit{e.g.,} differentiable BA or video-to-depth formulations~\cite{tang2018ba,teed2018deepv2d}, yet these still tend to degrade with distractor views common in web-sourced collections. Many pipelines~\cite{hong2024pf3plat,hong2024unifying,han2025d,wang2024dust3r} also rely on {COLMAP}~\cite{schonberger2016structure} for camera poses, supervision, or geometric verification during training/evaluation. More recent feed-forward correspondence/pointmap models~\cite{an2025cross,wang2025vggt,wang2024dust3r,leroy2024grounding} bypass iterative SfM modules and produce geometry directly, but they do not include explicit geometric verification. Our work complements this line by probing a feed-forward architecture, showing that internal layers encode signals correlated with outlier suppression, and leveraging them to select/reject views for more robust reconstruction under in-the-wild noise.\vspace{-5pt}

\paragrapht{Image retrieval and visual place recognition.}
Multi-view 3D reconstruction from web-sourced photo collections typically begins with keyword queries (e.g., “Rome”), followed by image retrieval or VPR to identify candidate views. Early approaches use bag-of-words~\cite{sivic2003video,nister2006scalable} or VLAD-based encoding~\cite{jegou2010aggregating}, while more recent systems adopt learned descriptors such as NetVLAD~\cite{arandjelovic2016netvlad}, GeM~\cite{radenovic2018fine}, MegaLoc~\cite{berton2025megaloc} and DELF~\cite{noh2017large}. These methods are widely used in retrieval-based SfM pipelines and are integrated into systems like COLMAP~\cite{schonberger2016structure} to initialize the view graph. However, retrieval similarity may not always reflect geometric suitability, and distractor images can persist. Our method provides a complementary mechanism by probing the internal layers of feed-forward 3D reconstruction models to identify and downweight distractors, improving robustness in uncurated image sets.

\section{Method}
\label{sec:method}
\begin{figure}
    \centering
    \includegraphics[width=1.0\linewidth]{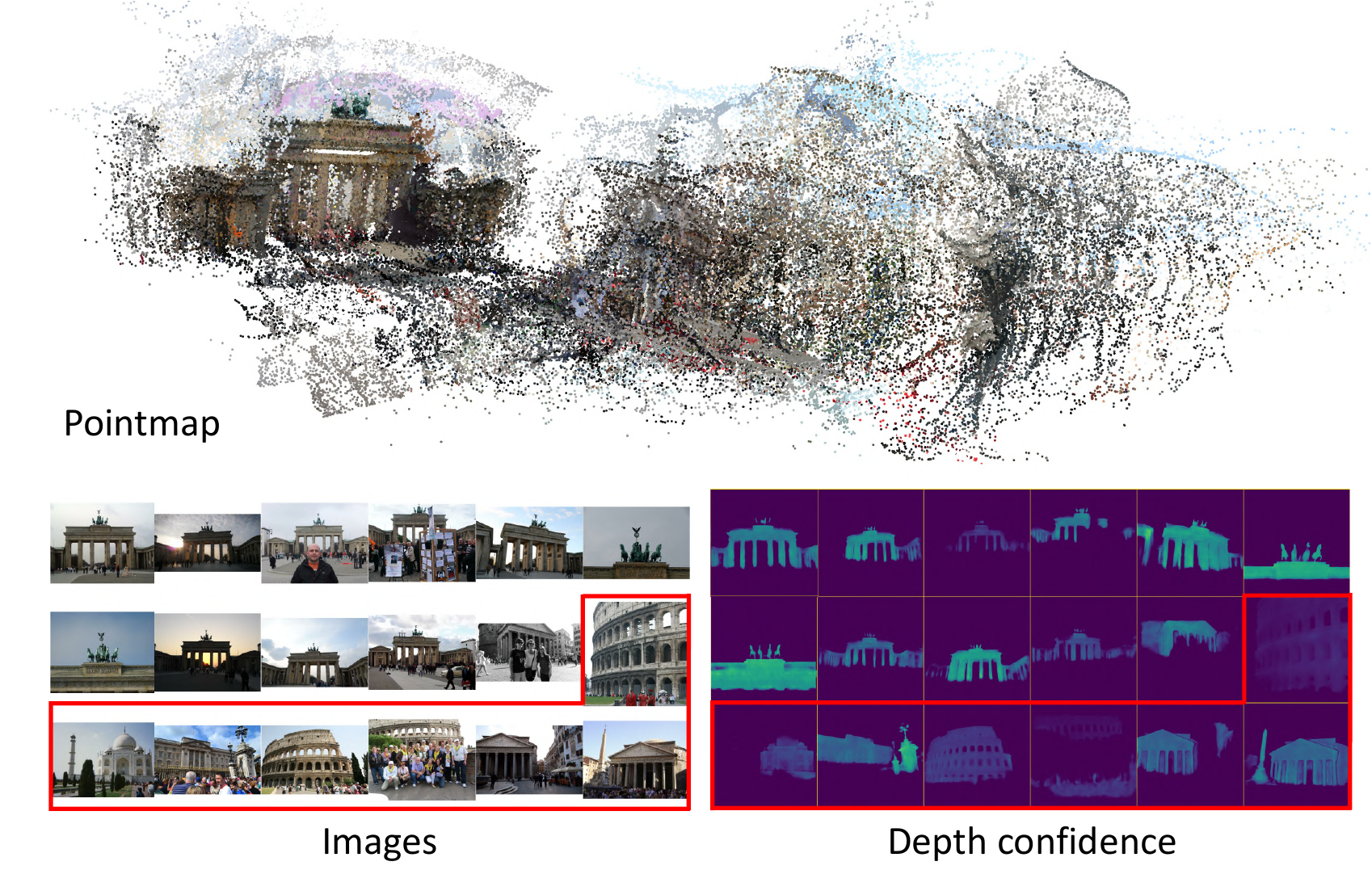}\vspace{-5pt}
    \caption{\textbf{Reconstruction by VGGT~\cite{wang2025vggt}.} Although VGGT predicts per-pixel confidence maps to down-weight unreliable depths, this post-hoc signal operates only at the point level and does not filter views. Consequently, distractor images are still reconstructed, allowing spurious content to corrupt the recovered geometry.}\vspace{-5pt}
    \label{fig:confidence}
\end{figure}

\subsection{Preliminaries}
\label{sec:preliminaries}

Feed-forward 3D reconstruction models~\cite{wang2024dust3r,leroy2024grounding,wang2025vggt} aim to recover scene geometry and camera poses directly from a set of $N$ images $\{I_1, \ldots, I_N\}$ without relying on explicit feature matching or optimization. These models learn to predict either per-view depth and pose ${(\mathrm{D}_i, \mathcal{P}_i)}$ or pointmaps ${\mathrm{X}_i}$ by first extracting image features and then modeling cross-view relationships using attention-based modules. Geometry is recovered by either regressing $\mathrm{X}_i$ directly or unprojecting $\mathrm{D}_i$ using estimated poses $\mathcal{P}_i$, with per-point confidence $\mathrm{C}_i$ optionally predicted. A representative model is the VGGT~\cite{wang2025vggt}, which uses a transformer-based encoder and alternating attention layers that consist of frame-wise and global attention to feed to dedicated decoder heads to estimate $\mathcal{P}_i$ and $\mathrm{D}_i$, along with a dedicated head for direct $\mathrm{X}_i$ regression. VGGT is trained end-to-end using supervision signals including camera pose loss, depth regression, and pointmap confidence, and outputs ${\mathcal{P}_i, \mathrm{X}_i, \mathrm{C}_i}$ in a single pass of inference.

\subsection{Motivation and overview}
Feed-forward 3D reconstruction models, such as DUSt3R~\cite{wang2024dust3r}, VGGT~\cite{wang2025vggt}, and Pi3~\cite{wang2025pi}, directly estimate scene geometry and camera poses from unstructured image collections, offering an efficient alternative to traditional SfM pipelines~\cite{schonberger2016structure}. However, a critical limitation—often overlooked in practice and unexplored—is the absence of explicit mechanisms for filtering noisy or irrelevant input views. This becomes especially pronounced in practical settings, where image collections frequently include off-topic scenes, or occlusions. Without built-in outlier rejection stages, such as view-graph construction or geometric verification~\cite{arandjelovic2016netvlad,schonberger2016structure}, these models remain vulnerable to distractor views, leading to degraded pose and geometry (see Fig.~\ref{fig:noise}).
\begin{figure}[t]
    \centering
    \includegraphics[width=0.9\linewidth]{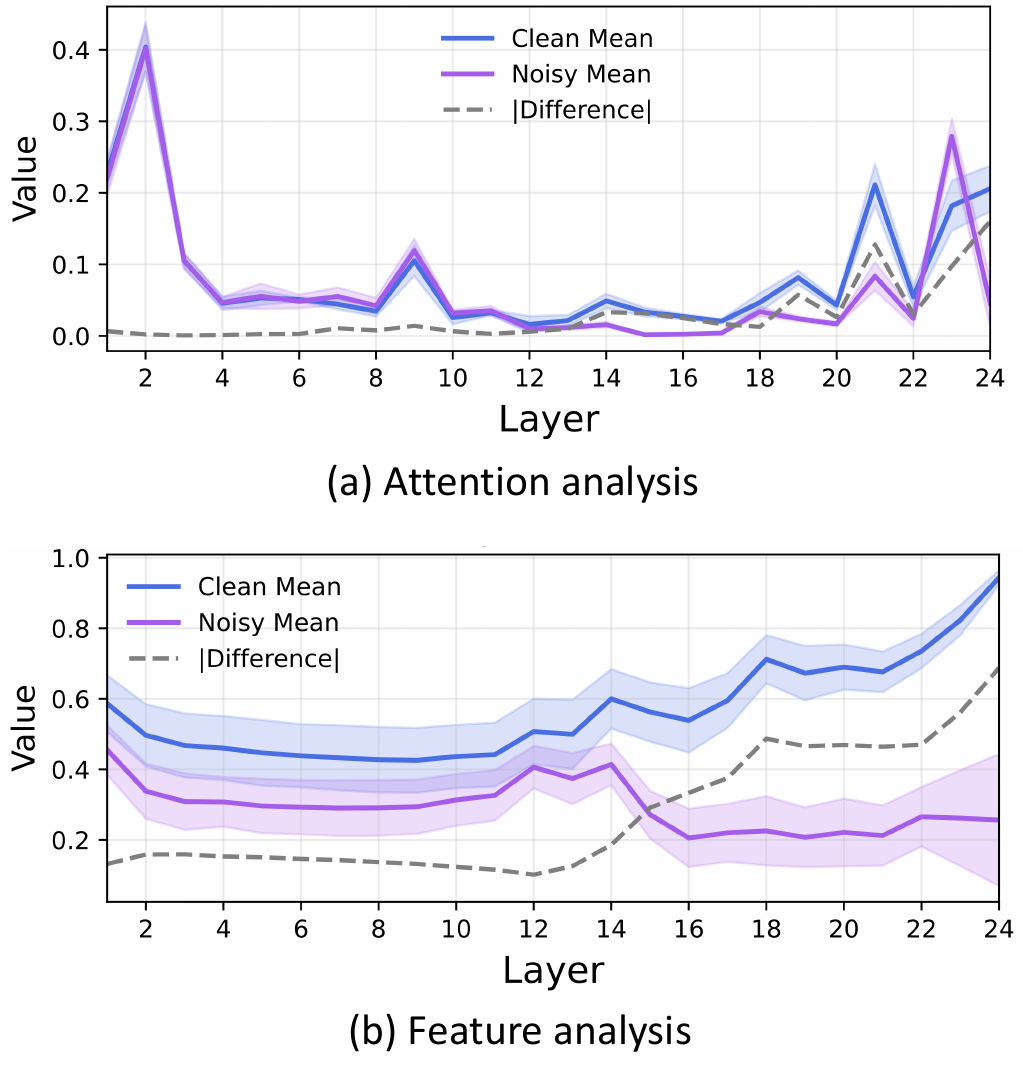}\vspace{-5pt}
  \caption{\textbf{Layer-wise analysis.} We measure the gap between clean and distractor views for attention and  feature similarity across VGGT’s all layers. The separation grows with depth and peaks at the final layer, indicating emergent noise suppression.}\vspace{-5pt}
    \label{fig:layer}
\end{figure}

\begin{figure*}[t]
    \centering
    \includegraphics[width=1\textwidth]{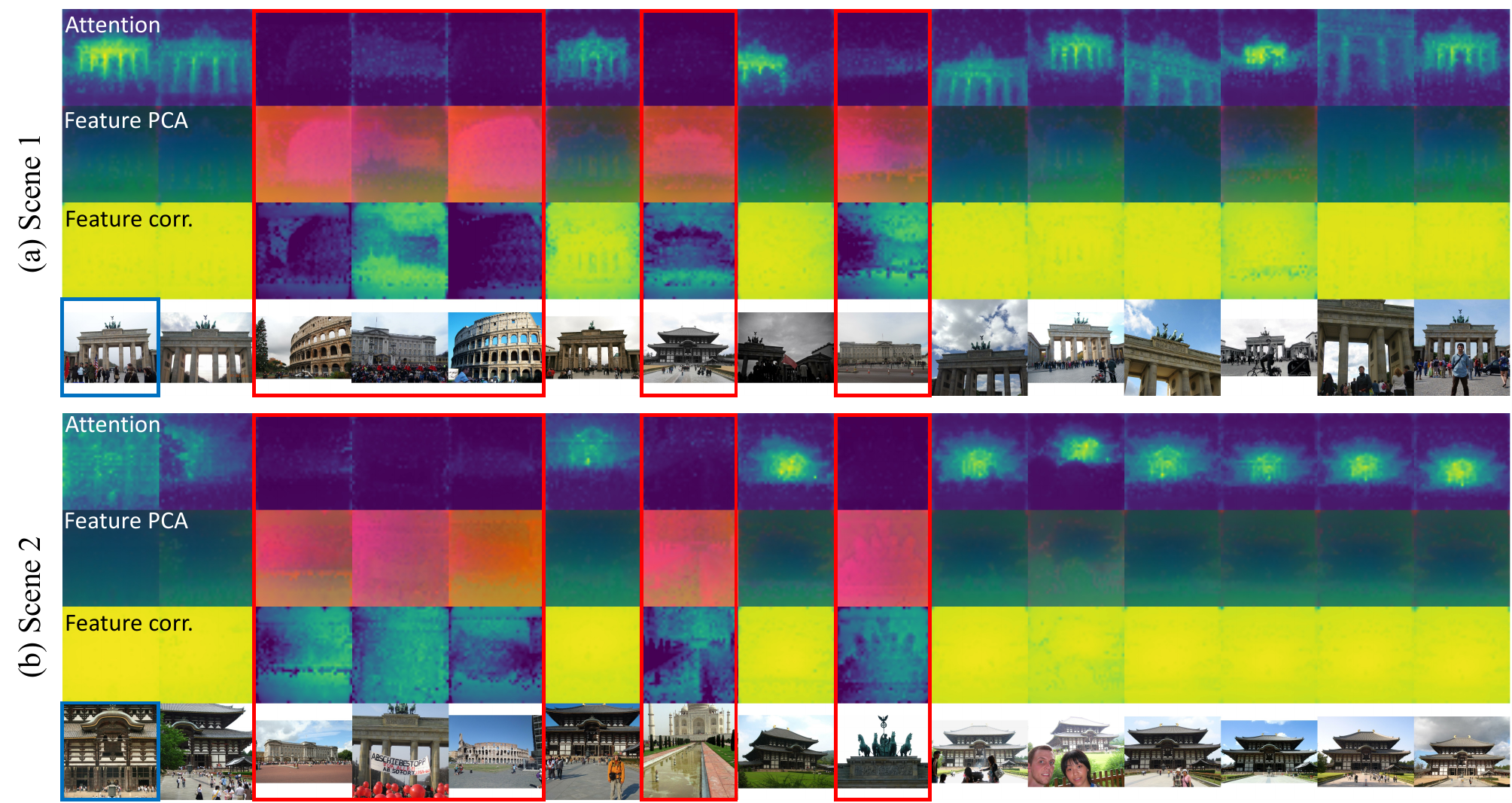}\vspace{-5pt}
  \caption{\textbf{Feature/attention visualization.} We show cross-view attention maps and intermediate feature–similarity maps from the final layer of VGGT on mixed sets containing clean and distractor images. Distractors are marked with \textcolor{red}{red boxes}; the query image is marked with a \textcolor{cvprblue}{blue box}. For each context image, scores are computed with respect to the query, averaged over all query tokens, and normalized for display. Both probes clearly downweight distractor views, revealing the model’s emergent view selectivity.}
\vspace{-5pt}
    \label{fig:feat_viz}
\end{figure*}

A common mitigation is to pre-filter with visual place recognition (VPR) or retrieval-based methods to prune distractors before reconstruction, following conventional SfM practice~\cite{schonberger2016structure}. While common, such approaches typically require per-scene hyperparameter tuning, and may not directly generalize across datasets or domains. Another seemingly reasonable option is to leverage the per-point confidence scores predicted with each pointmap to discard unreliable 3D points. Unfortunately, these confidences act \emph{post hoc} at the point level; without an explicit view-filtering mechanism, the system still reconstructs \emph{all} images, allowing distractor views to degrade poses and geometry, as shown in Fig.~\ref{fig:confidence}.  

In contrast, we observe \textit{an important emergent property}: VGGT~\cite{wang2025vggt}, despite being trained without any explicit outlier labels or filtering objectives, exhibits an emergent form of view selectivity in its internal representations. In particular, later-stage representations consistently downweight distractor views while emphasizing geometrically consistent ones. This insight enables a simple, training-free filtering strategy that leverages VGGT’s own attention/feature scores to rank and select views—eliminating additional modules, retraining, or hand-tuned retrieval pipelines—and yields robust scene geometry recovery from image collections consisting of noisy images.

The remainder of this section is structured as follows. In \textbf{\S~\ref{sec:problem}}, we formulate the view selection problem in the
context of multi-view feed-forward reconstruction.  In \textbf{\S~\ref{emergent}},  we begin with a pilot study to investigate the respective roles of feature  and attention maps and identify the key layer critical for view selection. Building on this, in \textbf{\S~\ref{sec:ir-view-filter}} we introduce internal representation-based view filtering approach.

\subsection{Problem formulation}
\label{sec:problem}

Given a set of $N$ uncalibrated images $\{I_1, I_2, \ldots, I_N\}$ capturing a static scene, the objective is twofold: (1) to estimate the 3D scene structure and camera parameters, and (2) to identify and discard noisy or irrelevant views that could degrade the reconstruction quality. Following the formulation of VGGT~\cite{wang2025vggt}, our model still predicts, for each image $I_i$, a camera pose $\mathcal{P}_i$, a per-pixel depth map $\mathrm{D}_i$, and a 3D pointmap $\mathrm{X}_i$, computed by lifting $\mathrm{D}_i$ into 3D space using $\mathcal{P}_i$. Additionally, the model outputs a confidence map $\mathrm{C}_i$ associated with each pointmap.

To enhance robustness to distractors, we aim to extract a clean subset of context images $\{I_j\}_{j \in \phi(i)} \subset \{I_1, \ldots, I_N\}$ for an arbitrary query image $I_i$. This is achieved by leveraging VGGT’s internal representations—intermediate feature maps $F_i$ and cross-view attention weights $A_{i\rightarrow j}$—where
$A_{i\rightarrow j}$ is obtained from dot-product attention between query and key projections. We define a selection function $\phi(i)$ that retains only the clean views $I_j$. The filtered context set $\{I_j\}_{j \in \phi(i)}$ is then \textit{re-fed} into the model to produce the final predictions $(\mathcal{P}_i, \mathrm{D}_i, \mathrm{X}_i)$ using only the most geometrically consistent views.

\subsection{Emergent noise suppression abilities}
\label{emergent}
We probe the internal representations of VGGT~\cite{wang2025vggt} to understand how it handles distractor views. To this end, we isolate the alternating attention stack and analyze, layer by layer, both the intermediate feature maps and the corresponding attention maps produced for a given image set.

\paragrapht{Deconstructing VGGT: Internal representations.}
Although feed-forward 3D reconstruction models like VGGT do not explicitly implement outlier rejection, we find that they exhibit \emph{emergent} filtering behavior in their internal representations, such as features and attention. Concretely, given a query image and a set of context images containing both clean and distracting views, we compute for each alternating-attention layer: (i) an attention score obtained by aggregating the weight that the query image that a user sets assigns to each context image, and (ii) the cosine similarity between the query feature map and each context feature map, which is computed pixel-wise on $\ell_2$-normalized features and then averaged over pixels to yield a single scalar per pair. We repeat this across layers and report, for each layer, the average scores for clean pairs and for distractor pairs, along with their gap (clean minus distractor), as shown in Fig.~\ref{fig:layer}.

From Fig.~\ref{fig:layer}(a), early layers show little separation between clean and distractor views; the gap grows steadily and peaks in the final layer, indicating that the model increasingly differentiates useful from irrelevant views as reasoning deepens. A parallel analysis on the feature maps reveals a similar trend with an even larger separation, as shown in Fig.~\ref{fig:layer}(b), suggesting that late-layer features act as a stronger discriminator of geometric relevance. Together, these findings support that the last layer serves as a critical gate that distinguishes views inconsistent with the query’s geometry. Notably, this behavior emerges without any supervision for outlier filtering, plausibly as a byproduct of optimizing for multi-view geometric consistency.\vspace{-5pt}

\paragrapht{Qualitative evidence.}
Fig.~\ref{fig:feat_viz} visualizes attention and feature responses, as well as the cosine similarity between feature responses, from the final layer. Irrelevant frames~(\textit{e.g.,} images from non-dominant scenes or with severe occlusions) receive low attention weight and weak feature similarity, whereas geometrically consistent views remain highly activated. These qualitative results align with the quantitative layer-wise trends and indicate that VGGT’s internal representations inherently distinguish distracting tokens and views. Motivated by this observation, our method leverages these late-layer similarity signals as a natural cue for identifying uninformative views.

\begin{figure}[t]
    \centering
    \includegraphics[width=1.0\linewidth]{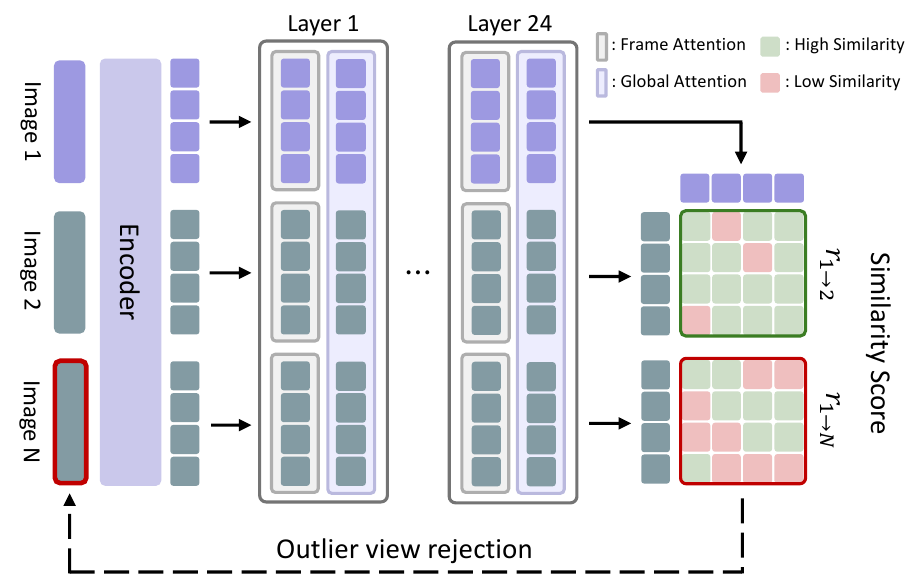}
 \caption{\textbf{Framework overview.} We compute per-view relevance from VGGT’s internal representations using two probes: (i) cross-view attention from query--key projections and (ii) cosine similarity of intermediate dense features. The resulting score $r_{i\rightarrow j}$ is thresholded with a single global $\tau$ to filter distractors, and the filtered set is re-fed to VGGT for reconstruction---without retraining or architectural changes.}
\vspace{-5pt}
    \label{fig:method_overview}
\end{figure}

\subsection{Outlier view rejection}
\label{sec:ir-view-filter}

We realize the selection function $\phi(i)$ using \emph{only} internal representations in VGGT~\cite{wang2025vggt}, acquired from a single feed-forward. As seen in the Fig.~\ref{fig:method_overview}, for a query (anchor) image $I_i$, we score each context image $I_j$ by aggregating cross-view attention or intermediate feature similarity on the final layer $\mathcal{L}$. 

\paragrapht{Attention score.}
Let $\ell^\star$ be the final attention layer and $A^{(\ell^\star)}$ the multi-head averaged attention. 
For a query (anchor) image $I_i$, the per-image attention score is the simple mean over tokens of $I_j$:
\begin{equation}
r^{\text{att}}_{i\rightarrow j}
\;=\;
\frac{1}{HW}\sum_{u,v}A^{(\ell^\star)}_{i\rightarrow j}(u,v),
\end{equation}
where $u,v$ indicate 2D spatial positions.
\paragrapht{Feature similarity score.}
Let $F^{(\ell^\star)}_i, F^{(\ell^\star)}_j \in \mathbb{R}^{H\times W\times d}$ be intermediate feature maps for images $I_i$ and $I_j$ from the same final layer. To measure feature-level similarity between two images, we first compute a pixel-wise correlation map~\cite{rocco2017convolutional} across all spatial positions of $l$-2 normalized feature maps $\tilde{F}^{(\ell^\star)}_i(u), \tilde{F}^{(\ell^\star)}_j(v)$:
\begin{equation}
{C}_{i\rightarrow j}(u,v)=\tilde{F}_i^{(\ell^\star)}(u)\cdot \tilde{F}_j^{(\ell^\star)}(v),
\end{equation}
where $u$ and $v$ index the $\tilde{F}_i$ and $\tilde{F}_j$. The overall feature similarity score is obtained by spatially averaging all correlations:
\begin{equation}
r^{\text{feat}}_{i\rightarrow j} =
\frac{1}{HW}
\sum_{u,v}
{C}_{i\rightarrow j}(u,v).
\end{equation}
This corresponds to computing the mean cosine similarity over the entire $HW\times HW$ correlation map.

\paragrapht{View rejection.}
Based on these, we introduce two strategies for view rejection: \textbf{RobustVGGT-$\mathcal A$}, which integrates attention-based scores, and \textbf{RobustVGGT-$\mathcal F$}, which utilizes feature similarity scores. The context set for $I_i$ is then
\begin{equation}
\phi(i) \;=\; \{\, j \;|\; j=i \;\text{ or }\; r_{i\rightarrow j}^{O}\ge\tau^O \,\},
\end{equation}
with $O \in \{\mathrm{att},\mathrm{feat}\}$ and a single fixed $\tau^O$ shared across all benchmarks. We finally re-run the backbone on $\{I_j\}_{j\in\phi(i)}$ to obtain $(\mathcal{P}_i,\mathrm{D}_i,\mathrm{X}_i,\mathrm{C}_i)$ using only spatially and geometrically consistent views.



\section{Experiments}
\label{sec:experiment}

\begin{figure*}[t]
    \centering
    \includegraphics[width=1\linewidth]{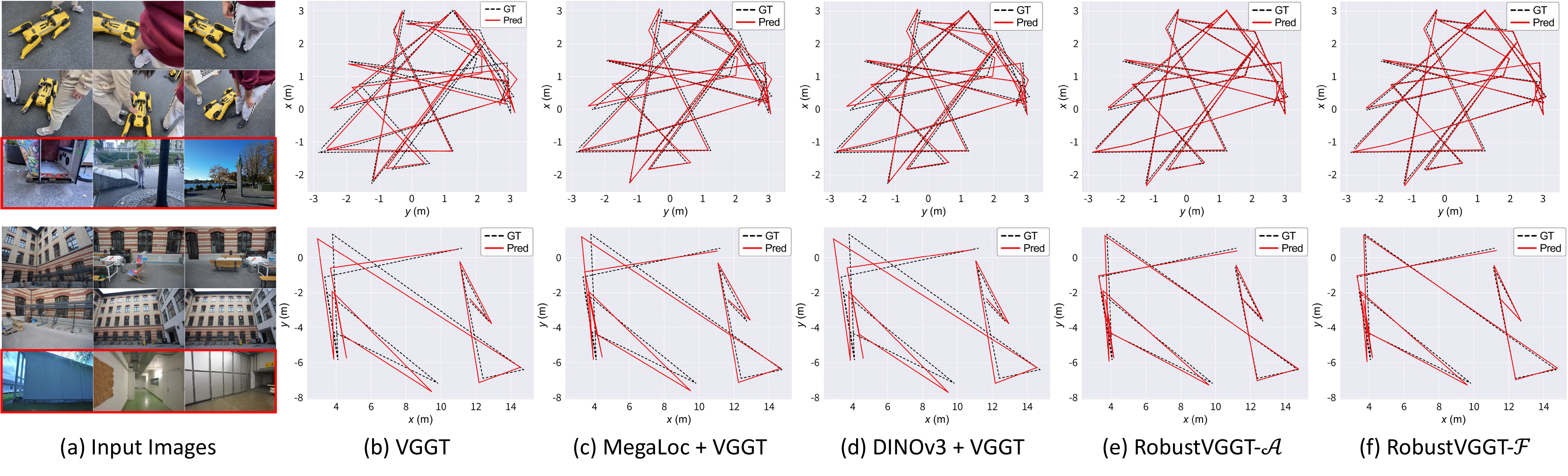}\vspace{-5pt}
    \caption{\textbf{Qualitative results of camera trajectory prediction.} Best viewed when zoomed in.}\vspace{-5pt}
    \label{fig:qual_camera}
\end{figure*}

\begin{figure*}[t]
    \centering
    \includegraphics[width=1\linewidth]{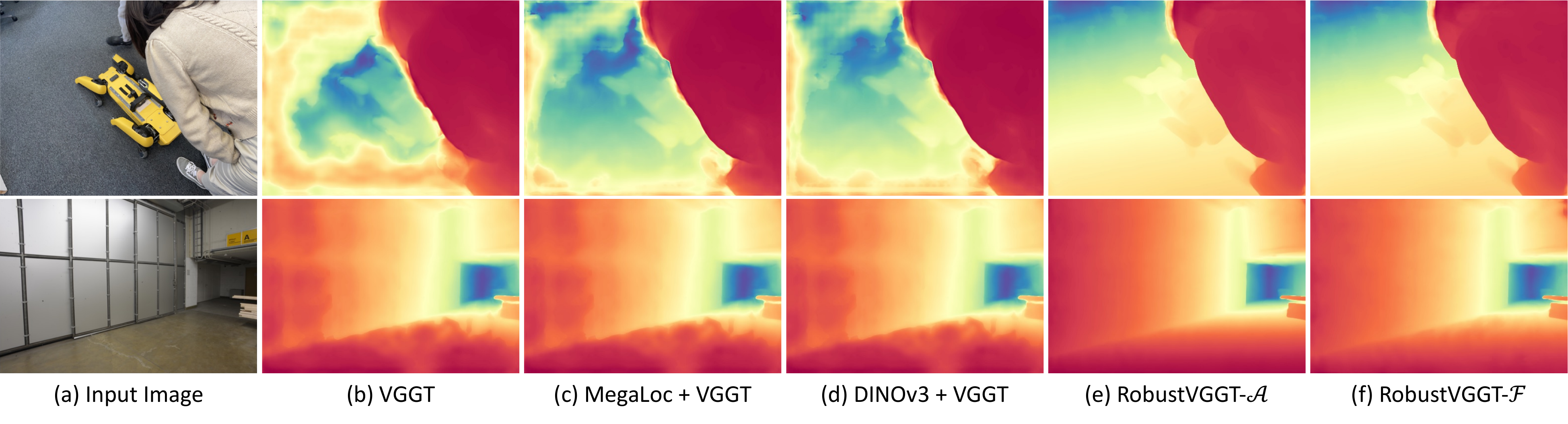}\vspace{-5pt}
    \caption{\textbf{Qualitative results of multi-view depth estimation.}}\vspace{-5pt}
    \label{fig:qual_depth}
\end{figure*}

\begin{table*}[t]
    \centering
  \caption{\textbf{Camera pose estimation across noise levels}. 
\textasteriskcentered\ denotes per-dataset hyperparameter tuning with \emph{oracle knowledge} of the number of clean images in the test set; these entries are \textcolor{gray}{shaded} as they are not directly comparable. 
$\dagger$ uses DINOv2 features extracted from VGGT.}
\vspace{-5pt}
    \label{tab:pose_eval}
    \resizebox{\textwidth}{!}{
    \begin{tabular}{l|ccc|ccc|ccc|ccc}
        \toprule
        \multirow{2}{*}{Methods} &
        \multicolumn{3}{c|}{Small} &
        \multicolumn{3}{c|}{Medium} &
        \multicolumn{3}{c|}{Large} &
        \multicolumn{3}{c}{Avg} \\
        & ATE $\downarrow$& RPE$_\text{trans}$ $\downarrow$& RPE$_\text{rot}$ $\downarrow$& ATE $\downarrow$& RPE$_\text{trans}$ $\downarrow$& RPE$_\text{rot}$ $\downarrow$& ATE $\downarrow$& RPE$_\text{trans}$ $\downarrow$& RPE$_\text{rot}$ $\downarrow$& ATE $\downarrow$& RPE$_\text{trans}$ $\downarrow$& RPE$_\text{rot}$ $\downarrow$\\
        \midrule
        \rowcolor{gray!20} \multicolumn{13}{c}{\textbf{Phototourism}} \\
        \midrule
        MASt3R-SfM~\cite{duisterhof2025mast3r} &   1.3556&   2.4084&   12.2887&   1.2683&   2.4735&   11.9315&   1.2329&   2.3143&   11.2861&   1.2856&   2.3987&   11.8354\\
        VGGT~\cite{wang2025vggt} & 0.3068& 0.4553& 0.9906&  0.3612&  0.5314&  1.1987&  0.3833&  0.5649&  1.3304&  0.3504&  0.5172&  1.1732\\
        \midrule
        MegaLoc~\cite{berton2025megaloc}+VGGT~\cite{wang2025vggt} &   0.2735&   \cellcolor{cvprblue!15}0.4076&   0.8841&   0.2999&   0.4460&   0.9872&   0.3161&   0.4700&   1.0714&   0.2965&   0.4412&   0.9809\\
        \textcolor{gray}{MegaLoc*~\cite{berton2025megaloc}+VGGT~\cite{wang2025vggt}} &   \textcolor{gray}{0.2689}&   \textcolor{gray}{0.4013}&   \textcolor{gray}{0.8686}&   \textcolor{gray}{0.2873}&   \textcolor{gray}{0.4274}&   \textcolor{gray}{0.9316}&   \textcolor{gray}{0.2997}&   \textcolor{gray}{0.4458}&   \textcolor{gray}{1.0020}&   \textcolor{gray}{0.2853}&   \textcolor{gray}{0.4248}&   \textcolor{gray}{0.9341}\\
        DINOv3~\cite{simeoni2025dinov3}+VGGT~\cite{wang2025vggt}&   0.3068&   0.4553&   0.9909&   0.3612&   0.5315&   1.1989&   0.3833&   0.5649&   1.3307&  0.3504&   0.5172&   1.1735\\
        DINOv2$^\dagger$~\cite{oquab2023dinov2}+VGGT~\cite{wang2025vggt}&   0.3068&   0.4553&   0.9906&   0.3612&   0.5314&   1.1987&   0.3833&   0.5649&   1.3304&  0.3504&   0.5172&   1.1732\\
        \midrule
         RobustVGGT-$\mathcal A$ & \cellcolor{cvprblue!15}0.2732& 0.4094&\cellcolor{cvprblue!15}0.8719& \cellcolor{cvprblue!15}0.2792&  \cellcolor{cvprblue!15}0.4153& \cellcolor{cvprblue!15}0.8806& \cellcolor{cvprblue!15}0.2930& \cellcolor{cvprblue!15}0.4349& \cellcolor{cvprblue!15}0.9310& \cellcolor{cvprblue!15}0.2818& \cellcolor{cvprblue!15}0.4199& \cellcolor{cvprblue!15}0.8945\\
         RobustVGGT-$\mathcal F$& \cellcolor{cvprblue!35}0.2641& \cellcolor{cvprblue!35}0.3936& \cellcolor{cvprblue!35}0.8420& \cellcolor{cvprblue!35}0.2645& \cellcolor{cvprblue!35}0.3949& \cellcolor{cvprblue!35}0.8402& \cellcolor{cvprblue!35}0.2664& \cellcolor{cvprblue!35}0.3973& \cellcolor{cvprblue!35}0.8388& \cellcolor{cvprblue!35}0.2650& \cellcolor{cvprblue!35}0.3953&\cellcolor{cvprblue!35}0.8403\\ 
        \midrule
        \rowcolor{gray!20} \multicolumn{13}{c}{\textbf{On-the-Go}} \\
        \midrule
        MASt3R-SfM~\cite{duisterhof2025mast3r} &   0.0754&   0.1548&   2.0281&   0.0749&   0.1474&   1.9934&   0.0768&   0.1510&   2.0328&   0.0757&   0.15107&   2.0181\\
        VGGT~\cite{wang2025vggt} & 0.0788& 0.1239& 1.0261&  0.1281&  0.1963&  1.4194&  0.1562&  0.2393&  1.7315&  0.1210&  0.1865&  1.3923\\
        \midrule
        MegaLoc~\cite{berton2025megaloc}+VGGT~\cite{wang2025vggt} &   0.0658&   0.1056&   0.9090&   0.1044&   0.1612&   1.1516&   0.1229&   0.1881&   1.3315&   0.0977&   0.1516&   1.1307\\
        \textcolor{gray}{MegaLoc*~\cite{berton2025megaloc}+VGGT~\cite{wang2025vggt}} &   \textcolor{gray}{0.0529}&   \textcolor{gray}{0.0872}&   \textcolor{gray}{0.8366}&   \textcolor{gray}{0.0580}&   \textcolor{gray}{0.0942}&   \textcolor{gray}{0.8812}&   \textcolor{gray}{0.0637}&   \textcolor{gray}{0.1033}&   \textcolor{gray}{0.9464}&   \textcolor{gray}{0.0582}&   \textcolor{gray}{0.0949}&   \textcolor{gray}{0.8881}\\
        DINOv3~\cite{simeoni2025dinov3}+VGGT~\cite{wang2025vggt}&   0.0675&   0.1077&   0.9184&   0.1075&   0.1667&  \cellcolor{cvprblue!15} 1.2201&   0.1312&   0.2025&   1.4602&   0.1021&   0.1589&   1.1996\\
        DINOv2$^\dagger$~\cite{oquab2023dinov2}+VGGT~\cite{wang2025vggt}&   0.0788&   0.1239&   1.0261&   0.1281&   0.1963&   1.4194&   0.1562&   0.2393&   1.7315&  0.1210&   0.1865&   1.3923\\
        \midrule
         RobustVGGT-$\mathcal A$ &\cellcolor{cvprblue!15} 0.0578& \cellcolor{cvprblue!15}0.0952&\cellcolor{cvprblue!15}0.8800& \cellcolor{cvprblue!15}0.0790& \cellcolor{cvprblue!15}0.1253&  1.2922& \cellcolor{cvprblue!15}0.0697& \cellcolor{cvprblue!15}0.1126& \cellcolor{cvprblue!15}1.0361& \cellcolor{cvprblue!15}0.0688& \cellcolor{cvprblue!15}0.1110& \cellcolor{cvprblue!15}1.0694\\
         RobustVGGT-$\mathcal F$& \cellcolor{cvprblue!35}0.0521& \cellcolor{cvprblue!35}0.0861& \cellcolor{cvprblue!35}0.8179& \cellcolor{cvprblue!35}0.0568& \cellcolor{cvprblue!35}0.0931& \cellcolor{cvprblue!35}0.8914& \cellcolor{cvprblue!35}0.0660&\cellcolor{cvprblue!35} 0.1055&\cellcolor{cvprblue!35} 0.9872& \cellcolor{cvprblue!35}0.0583& \cellcolor{cvprblue!35}0.0949&\cellcolor{cvprblue!35}0.8988\\ 
        \midrule
        \rowcolor{gray!20} \multicolumn{13}{c}{\textbf{RobustNeRF}} \\
        \midrule
        MASt3R-SfM~\cite{duisterhof2025mast3r} & \cellcolor{cvprblue!35}0.1153 &  0.2597 &2.7124  &\cellcolor{cvprblue!35}0.1196  & 0.2650  & 2.7617 & \cellcolor{cvprblue!35}0.1182  & 0.2626 & 2.6882 &  \cellcolor{cvprblue!35}0.1177& 0.2624& 2.7208\\
        VGGT~\cite{wang2025vggt} & 0.1519& 0.2742& 1.2052&  0.1598&  0.2908&  1.2311&  0.1680&  0.3062&  1.3493&  0.1599&  0.2904&  1.2619\\
        \midrule
        MegaLoc~\cite{berton2025megaloc}+VGGT~\cite{wang2025vggt} &   0.1496&   0.2692&   1.1920&   0.1573&   0.2847&   1.2199&   0.1618&   0.2934&   1.2721&   0.1562&   0.2824&   1.2280\\
        \textcolor{gray}{MegaLoc*~\cite{berton2025megaloc}+VGGT~\cite{wang2025vggt}} &   \textcolor{gray}{0.1352}&   \textcolor{gray}{0.2418}&   \textcolor{gray}{1.1707}&   \textcolor{gray}{0.1358}&  \textcolor{gray}{0.2416}&  \textcolor{gray}{1.1663}&   \textcolor{gray}{0.1356}&   \textcolor{gray}{0.2394}&  \textcolor{gray}{1.1552}&   \textcolor{gray}{0.1355}&   \textcolor{gray}{0.2409}&   \textcolor{gray}{1.1641}\\
        DINOv3~\cite{simeoni2025dinov3}+VGGT~\cite{wang2025vggt}&   0.1516&   0.2738&   1.2032&   0.1595&   0.2902&   1.2296&   0.1677&   0.3055&   1.3426&   0.1596&   0.2898&   1.2585\\
        DINOv2$^\dagger$~\cite{oquab2023dinov2}+VGGT~\cite{wang2025vggt}&   0.1519&   0.2742&   1.2052&   0.1598&   0.2908&   1.2311&   0.1680&   0.3062&   1.3493&  0.1599&   0.2904&   1.2619\\
        \midrule
         RobustVGGT-$\mathcal A$ & \cellcolor{cvprblue!15}0.1361&\cellcolor{cvprblue!35}0.2433& \cellcolor{cvprblue!15}1.1766&  0.1379&  \cellcolor{cvprblue!15}0.2447&  \cellcolor{cvprblue!35}1.1657&  0.1406& \cellcolor{cvprblue!15}0.2478&  \cellcolor{cvprblue!15}1.1632&  0.1382&  \cellcolor{cvprblue!15}0.2453&  \cellcolor{cvprblue!15}1.1685\\
         RobustVGGT-$\mathcal F$& 0.1388& \cellcolor{cvprblue!15}0.2480& \cellcolor{cvprblue!35}1.1656&\cellcolor{cvprblue!15}0.1374&\cellcolor{cvprblue!35}0.2432& \cellcolor{cvprblue!15}1.1670&\cellcolor{cvprblue!15}0.1374& \cellcolor{cvprblue!35}0.2415& \cellcolor{cvprblue!35}1.1514& \cellcolor{cvprblue!15}0.1379& \cellcolor{cvprblue!35}0.2442&\cellcolor{cvprblue!35}1.1613\\ 
        \midrule
        \rowcolor{gray!20} \multicolumn{13}{c}{\textbf{ETH3D}} \\
        \midrule
        MASt3R-SfM~\cite{duisterhof2025mast3r} &   2.3871&  4.2586 &  77.368 &  2.3931 &4.2766   &  78.521 &  2.3796 &  4.1958 &  76.4021 & 2.3866  &  4.2437 & 77.4303  \\
        VGGT~\cite{wang2025vggt} & 0.8572& 1.3675& 6.3908&  0.9182&  1.5028&   9.6272&  1.0165&  1.7348&  15.2774&  0.9306&  1.5350&  10.4318\\
        \midrule
        MegaLoc~\cite{berton2025megaloc}+VGGT~\cite{wang2025vggt} &   0.9275&   1.4470&   5.5320&   0.9233&   1.4789&   7.5607&   0.9639&   1.5977&   11.4364&   0.9382&   1.5079&   8.1764\\
        \textcolor{gray}{MegaLoc*~\cite{berton2025megaloc}+VGGT~\cite{wang2025vggt} }&\textcolor{gray}{0.9170}&   \textcolor{gray}{1.4904}&   \textcolor{gray}{3.7593}&   \textcolor{gray}{0.9418}&   \textcolor{gray}{1.5194}&   \textcolor{gray}{4.3126}&   \textcolor{gray}{0.9800}&   \textcolor{gray}{1.5802}&   \textcolor{gray}{5.7108}&   \textcolor{gray}{0.9463}&   \textcolor{gray}{1.5300}&   \textcolor{gray}{4.5942}\\
        DINOv3~\cite{simeoni2025dinov3}+VGGT~\cite{wang2025vggt}&   0.8551&   1.3667&   6.2305&   0.9113&   1.4891&   9.4103&   1.0027&   1.7022&   14.9265&   0.9230&   1.5193&   10.1891\\
        DINOv2$^\dagger$~\cite{oquab2023dinov2}+VGGT~\cite{wang2025vggt}&   0.8556&   1.3661&   6.2515&   0.9188&   1.5022&   9.5374&   1.0165&   1.7346&   15.2582&  0.9303&   1.5343&   10.3490\\
        \midrule
         RobustVGGT-$\mathcal A$ & \cellcolor{cvprblue!15}0.7447& \cellcolor{cvprblue!15}1.1724& \cellcolor{cvprblue!15}3.8938&  \cellcolor{cvprblue!35}0.7673&  \cellcolor{cvprblue!35}1.2123& \cellcolor{cvprblue!15}3.7325& \cellcolor{cvprblue!35}0.6874&  \cellcolor{cvprblue!35}1.0708& \cellcolor{cvprblue!15}4.4779& \cellcolor{cvprblue!35}0.7331&  \cellcolor{cvprblue!35}1.1518&  \cellcolor{cvprblue!15}4.0347\\
         RobustVGGT-$\mathcal F$&\cellcolor{cvprblue!35}0.6224& \cellcolor{cvprblue!35}1.0300& \cellcolor{cvprblue!35}2.7304&\cellcolor{cvprblue!15}0.8038& \cellcolor{cvprblue!15}1.3159& \cellcolor{cvprblue!35}2.9959& \cellcolor{cvprblue!15}0.8636& \cellcolor{cvprblue!15}1.3882& \cellcolor{cvprblue!35}3.4866& \cellcolor{cvprblue!15}0.7633&\cellcolor{cvprblue!15}1.2447 & \cellcolor{cvprblue!35}3.0710\\ 
        \bottomrule
    \end{tabular}}\vspace{-5pt}
\end{table*}

\begin{table*}[t]
    \centering
  \caption{\textbf{Multi-view depth estimation results}. Gray text indicates methods not directly comparable; they are included for reference only.}\vspace{-5pt}

    \label{tab:mvs_benchmark_results}
    \resizebox{0.8\linewidth}{!}{
    \begin{tabular}{l|cccccccc}
        \toprule
        \multirow{2}{*}{Methods} &
        \multicolumn{8}{c}{\cellcolor{gray!20}ETH3D} \\
        & \multicolumn{2}{c}{Small} & \multicolumn{2}{c}{Medium} & \multicolumn{2}{c}{Large} & \multicolumn{2}{c}{Avg}  \\
        & AbsRel $\downarrow$ & $\delta<1.25 \uparrow$ 
        & AbsRel $\downarrow$ & $\delta<1.25 \uparrow$
        & AbsRel $\downarrow$ & $\delta<1.25 \uparrow$
        & AbsRel $\downarrow$ & $\delta<1.25 \uparrow$ \\
        \midrule
        MASt3R-SfM~\cite{duisterhof2025mast3r} & 0.0779& 0.9329& 0.0789& 0.9321& 0.0777& 0.9327& 0.0782& 0.9326\\
        VGGT~\cite{wang2025vggt} & 0.0431& 0.9683& 0.0422& 0.9730& 0.0403& 0.9776& 0.0419& 0.9730 \\
        \midrule
        MegaLoc~\cite{berton2025megaloc}+VGGT~\cite{wang2025vggt} & 0.0391& 0.9729& 0.0435& 0.9694& 0.0409& 0.9759&0.0411 & 0.9727 \\
        \textcolor{gray}{MegaLoc*~\cite{berton2025megaloc}+VGGT~\cite{wang2025vggt}} &  \textcolor{gray}{ 0.0291}&  \textcolor{gray}{ 0.9844}&  \textcolor{gray}{ 0.0318}&   \textcolor{gray}{0.9822}&   \textcolor{gray}{0.0365}&   \textcolor{gray}{0.9772}&   \textcolor{gray}{0.0325}&  \textcolor{gray}{0.9813}\\
        DINOv3~\cite{simeoni2025dinov3}+VGGT~\cite{wang2025vggt}&   0.0430&   0.9683&   0.0422&   0.9729&   0.0399&   0.9780&   0.0417&   0.9731\\
        DINOv2$^\dagger$~\cite{oquab2023dinov2}+VGGT~\cite{wang2025vggt}&   0.0431&   0.9683&   0.0422&   0.9730&   0.0403&   0.9776&   0.0419&    0.9730\\
        \midrule
         RobustVGGT-$\mathcal A$ & \cellcolor{cvprblue!15}0.0308& \cellcolor{cvprblue!15}0.9828& \cellcolor{cvprblue!15}0.0313& \cellcolor{cvprblue!15}0.9822& \cellcolor{cvprblue!15}0.0351& \cellcolor{cvprblue!15}0.9770& \cellcolor{cvprblue!15}0.0324 &\cellcolor{cvprblue!15}0.9807  \\
         RobustVGGT-$\mathcal F$ &  \cellcolor{cvprblue!35}0.0288 & \cellcolor{cvprblue!35}0.9840 &  \cellcolor{cvprblue!35}0.0297 & \cellcolor{cvprblue!35}0.9833  & \cellcolor{cvprblue!35}0.0319  &  \cellcolor{cvprblue!35}0.9813 & \cellcolor{cvprblue!35}0.0301 &\cellcolor{cvprblue!35}0.9829\\ 
        \bottomrule
    \end{tabular}}\vspace{-5pt}
\end{table*}


\subsection{Experimental settings}
\paragraph{Datasets.}
We evaluate across four benchmarks to cover web-sourced noise, casual captures and explicit distractors. Phototourism~\cite{jin2021image} comprises Internet-sourced image collections with large appearance changes (lighting, weather) and transient objects; we use six landmark scenes: Brandenburg Gate, Buckingham Palace, Colosseum, Pantheon, Taj Mahal, and Temple Naraa to assess robustness under real-world variability.
On-the-go~\cite{ren2024nerf} contains casually captured indoor/outdoor sequences with motion blur, occlusions, and uneven viewpoint coverage; we evaluate nine sequences: corner, drone, fountain, mountain, patio, patio\_high, spot, statue, and train.
RobustNeRF~\cite{sabour2023robustnerf} augments scenes with distractor images, providing a controlled testbed for outlier rejection.
Following prior work~\cite{wang2025vggt}, we also report results on ETH3D~\cite{schops2017multi}, a high-quality multi-view benchmark with accurate ground-truth poses and depth for assessing geometric precision. 

\paragrapht{Task and baselines.}
We assess our method on two downstream tasks—multi-view pose estimation and multi-view depth estimation—which jointly gauge the fidelity of the recovered 3D scene. We compare against VGGT~\cite{wang2025vggt} and  against MASt3R-SfM~\cite{duisterhof2025mast3r}, which reconstructs by building a view/scene graph from MASt3R encoder features and optimizing over it.
In addition, we instantiate three VGGT-based pre-filtering baselines that apply view selection \emph{before} running VGGT:
(i) MegaLoc+VGGT uses an off-the-shelf visual place recognition model~\cite{berton2025megaloc} to filter distracting images and passes the resulting context set $\phi(i)$ to VGGT;
(ii) DINOv3+VGGT replaces VPR with DINOv3~\cite{simeoni2025dinov3} global descriptors to select views;
(iii) DINOv2$^\dagger$+VGGT leverages the VGGT encoder’s DINO-v2~\cite{oquab2023dinov2} features for view selection.

\paragrapht{Evaluation protocol.}
To our knowledge, this is the first work to evaluate feed-forward reconstruction under controlled \emph{view-noise} levels, so we introduce a simple, reproducible protocol with three settings, \textit{Small}, \textit{Medium}, and \textit{Large}, that differ only in the number of distractor images. For each trial, we sample $N_c=30$ \textbf{clean} images from the same scene and $N_n\in\{10,30,50\}$ \textbf{distracting} images uniformly from \emph{other} scenes, yielding sets of size $N_c{+}N_n\in\{40,60,80\}$. The sampling is repeated $10$ times with different seeds, and we report mean over trials. All methods are evaluated on the \emph{identical} sampled sets for fairness. For ETH3D~\cite{schops2017multi}, some scenes contain fewer images; we therefore set $N_c{=}14$ and use $N_n\in\{5,14,30\}$ for the Small/Medium/Large settings, respectively. In all datasets, clean and distacting pools are disjoint, and selection is scene-agnostic, which means no metadata or geometry is used. The same protocol is used for both evaluation tasks, multi-view pose and depth, and the noise level is the only factor varied across settings.

\subsection{Experimental results}
\paragraph{Camera pose estimation.}
We evaluate trajectories using three standard metrics: 
ATE, a Sim(3)-aligned RMSE of camera centers, 
RPE$_\text{trans}$ and RPE$_\text{rot}$, translational/rotational errors between consecutive frames. 
We report mean metrics over 10 random trials per noise level.

\begin{figure*}[t]
    \centering
    \includegraphics[width=0.95\linewidth]{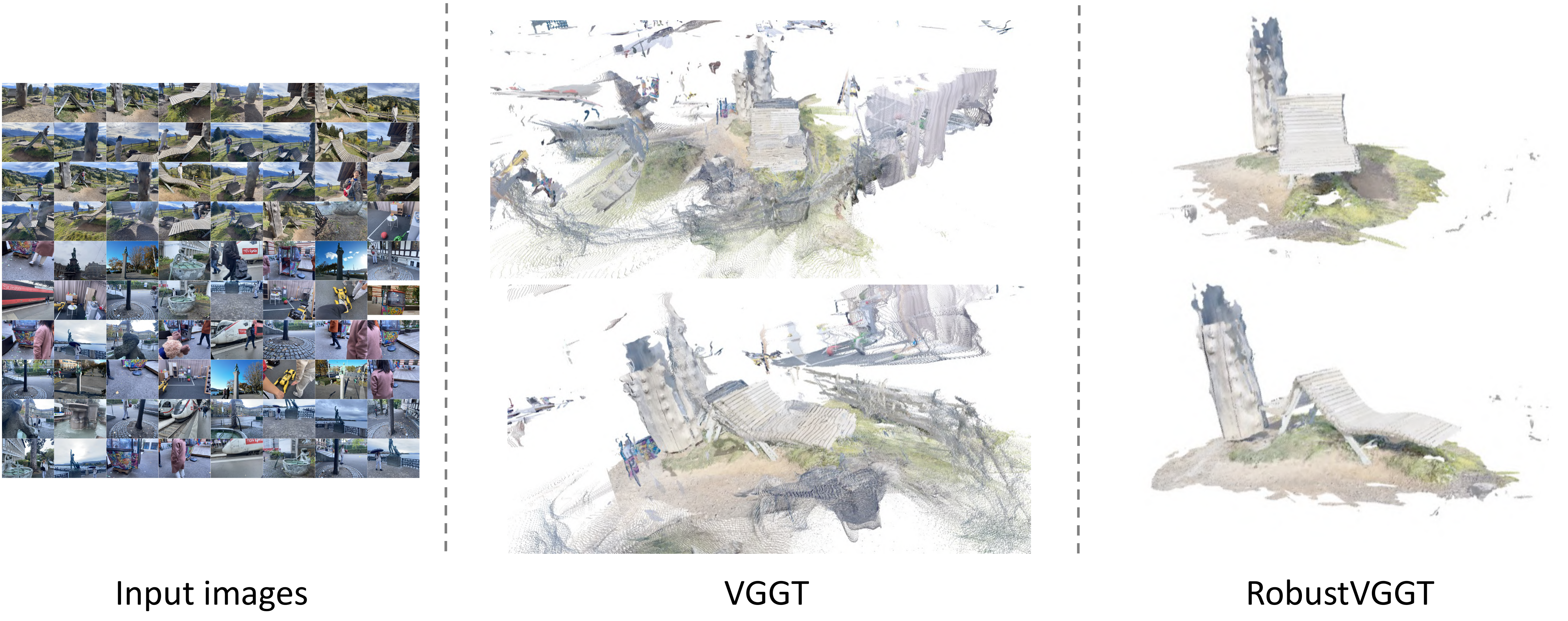}
    \vspace{-5pt}
    \caption{\textbf{Visualization of point maps produced by VGGT and RobustVGGT-$\mathcal{F}$.}}
    \label{fig:pointmap_data}
\end{figure*}

\begin{figure*}[t]
    \centering
    \includegraphics[width=0.95\linewidth]{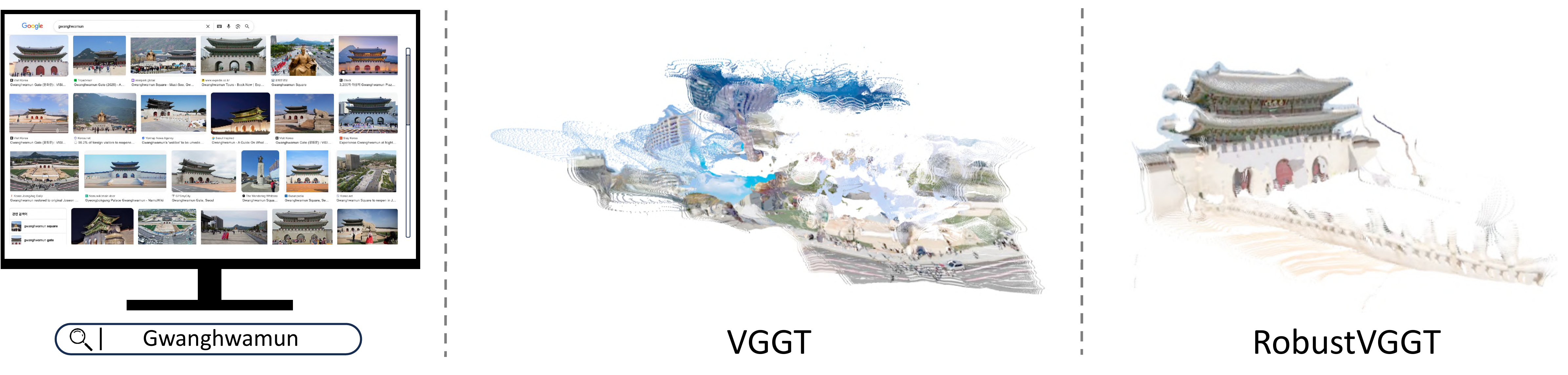}
    \vspace{-5pt}
    \caption{\textbf{Visualization of point maps produced by VGGT and RobustVGGT-$\mathcal{F}$ on internet-collected images.} } 
    \label{fig:pointmap_internet}
    \vspace{-10pt}
\end{figure*}

\begin{figure}[t]
    \centering
    \includegraphics[width=\linewidth]{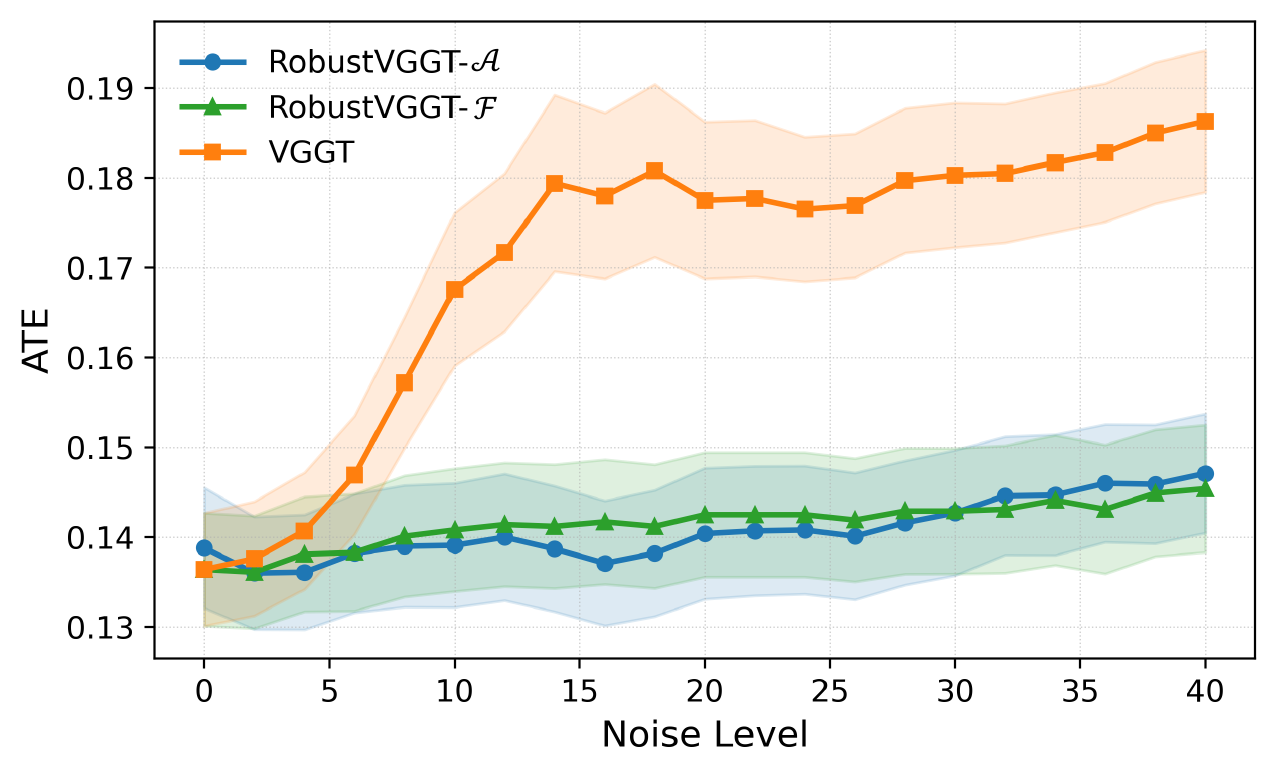}\vspace{-5pt}
\caption{\textbf{Camera ATE vs. noise level.} VGGT without explicit view filtering degrade as the number of noisy views increases; Our approach enables robust 3D reconstruction.  }\vspace{-5pt}

    \label{fig:noise}
\end{figure}

Quantitative results are summarized in Tab.~\ref{tab:pose_eval}, with representative trajectories in Fig.~\ref{fig:qual_camera}. 
\textit{(i) Baselines without explicit filtering.} VGGT shows a trend that performance degrades from \textit{Small}$\rightarrow$\textit{Medium}$\rightarrow$\textit{Large} noise, with ATE and RPE increasing as more distractors are introduced, reflecting the absence of a view-filtering mechanism. In contrast, ours shows more robust performance, without significant performance degradations. This trend is also shown in Fig.~\ref{fig:noise}. 
\textit{(ii) Pre-filtering baselines.} MegaLoc+VGGT, DINOv3+VGGT, and DINO$^{\dagger}$+VGGT mitigate some errors but despite their dataset-specific hyperparameter tuning, which is determined by providing an oracle knowledge of the number of clean images,  our method achieves the lowest errors  using a \emph{single global threshold}. Note that on RobustNeRF, MASt3R\textendash SfM achieves lower ATE than our method. We attribute this to the dataset’s small, highly overlapping scenes, where global view-graph optimization and bundle adjustment are particularly effective, diminishing the relative benefit of pre-filtering.
\textit{(iii) DINO-based variants.} Although DINO features are known for their strong generalization,  our method yields larger gains across all noise levels, suggesting that strong descriptors alone are insufficient without geometry-aware selection.

\vspace{-5pt}
\paragrapht{Multi-view depth estimation.}
Following practice in affine-invariant depth estimation~\cite{ke2024repurposing, yang2024depth}, we align the predicted depth maps to the ground truth using a scale and shift, and evaluate performance using AbsRel, a mean absolute relative error, and $\delta<1.25$, percentage of pixels whose predicted-to-GT depth ratio is within 1.25, on valid pixels after alignment of predicted and ground-truth depths. 
Tab.~\ref{tab:mvs_benchmark_results}  shows the quantitative comparisons that exhibit the similar trend as for poses: methods without explicit view filtering, \textit{e.g.,} VGGT and MASt3R\textendash SfM,  tend to degrade from when the distracting images are introduced. 
Pre-filtering baselines recover part of the loss but our variants, RobustVGGT-$\mathcal{A}$ and RobustVGGT-$\mathcal{F}$, achieve the best results across all noise levels. We also observe apparent differences in visualized multi-view depth, as exemplified in Fig.~\ref{fig:qual_depth}. We also present the point map visualizations of VGGT and our method in Fig.~\ref{fig:pointmap_data} and Fig.~\ref{fig:pointmap_internet}. As can be seen from the chair example in Fig.~\ref{fig:pointmap_data}, without noise filtering, the reconstruction for a clean image is noticeably degraded. In Fig.~\ref{fig:pointmap_internet}, we also show our robustness of reconstruction on internet-collected images.

\subsection{Ablation studies and analysis}
\paragraph{Effects of varying threshold hyperparameter.} In this ablation study, we investiage the effects of varying the threshold value for $\tau^O$. We use Phototourism and On-the-Go datasets to validate both the generalizability  and robustness of our approach. We include both of our variants, VGGT-$\mathcal{A}$ and VGGT-$\mathcal{F}$. The results are summarized in Tab.~\ref{tab:threshold_ablation}. From the results,  
for RobustVGGT-$\mathcal A$, the best camera estimation performance was achieved at $\tau=0.05$, while for RobustVGGT-$\mathcal F$, the optimal value was $\tau=0.65$. These threshold values were therefore used for all evaluations. 

\paragrapht{Success rate of distractor rejection.}
Tab.~\ref{tab:success_rate} shows that our method consistently rejects distractors more reliably than the baselines across datasets. For example, on On-the-Go and ETH3D, VGGT-$\mathcal{F}$ attains the highest average success and VGG-$\mathcal{A}$ achieves the highest success rate on other datasets. We attribute this to VGGT’s \emph{cross-view} reasoning, which provides stronger rejection signals than single-view global descriptors typical in VPR. Interestingly, DINOv3 performs poorly for distractor rejection: regardless of how we derive a global descriptor, \textit{e.g.,} CLS token or GeM pooling~\cite{radenovic2018fine}, or whether we use local-token similarities, the success rate remains low. We attribute this to its appearance/semantics-driven training—features group visually similar images but are largely agnostic to geometric overlap—so non-overlapping yet look-alike distractors are not reliably filtered, and the model is not trained for view selection.

\begin{table}[t]
\centering
\caption{\textbf{Ablation study on varying threshold values $\tau$.} We report the results of (a) RobustVGGT-$\mathcal A$ and (b) RobustVGGT-$\mathcal F$ on Phototourism and On-the-Go using camera pose estimation metrics. Results are formatted as Phototourism / On-the-Go.}\vspace{-5pt}
\begin{subtable}[t]{0.47\linewidth}
\centering
\resizebox{\linewidth}{!}{
\begin{tabular}{c|ccc}
\toprule
$\tau^{\text{attn}}$ & ATE & RPE$_\text{trans}$ & RPE$_\text{rot}$ \\
\midrule
0.05 & 0.2818 / 0.0688& 0.4199 / 0.1110& 0.8945 / 1.0694\\
0.04 & 0.2968 / 0.0729& 0.4403 / 0.1172& 0.9419 / 1.0266\\
0.03 & 0.3130 / 0.1046& 0.4632 / 0.1629& 1.0174 / 1.4162\\
0.02 & 0.3418 / 0.1132& 0.5039 / 0.1748& 1.1306 / 1.2880\\
\bottomrule
\end{tabular}}
\vspace{1pt}
\caption{RobustVGGT-$\mathcal A$}
\end{subtable}
\hfill
\begin{subtable}[t]{0.47\linewidth}
\centering
\resizebox{\linewidth}{!}{
\begin{tabular}{c|ccc}
\toprule
$\tau^{\text{feat}}$ & ATE & RPE$_\text{trans}$ & RPE$_\text{rot}$ \\
\midrule
0.65 & 0.2650 / 0.0583& 0.3953 / 0.0949& 0.8403 / 0.8988\\
0.55 & 0.2748 / 0.0671& 0.4088 / 0.1085& 0.8584 / 1.1270\\
0.45 & 0.2870 / 0.0702& 0.4254 / 0.1128& 0.8966 / 0.8966\\
0.35& 0.2777 / 0.0820& 0.4114 / 0.1303&0.9708 / 1.2868\\
\bottomrule
\end{tabular}}
\vspace{1pt}
\caption{RobustVGGT-$\mathcal F$}
\end{subtable}
\label{tab:threshold_ablation}
\vspace{-5pt}
\end{table}

\begin{table}[t]
\centering
\caption{\textbf{Success rate of distractor rejection across noise levels.}}
\vspace{-5pt}
\label{tab:success_rate}
\resizebox{\linewidth}{!}{
\begin{tabular}{l|l|ccccc}
\toprule
Dataset & Noise level & MegaLoc~\cite{berton2025megaloc}  &\textcolor{gray}{MegaLoc*~\cite{berton2025megaloc}}  &DINOv3~\cite{simeoni2025dinov3}&RobustVGGT-$\mathcal F$& RobustVGGT-$\mathcal A$ \\
\midrule
\multirow{4}{*}{Phototourism} 
 & Small  &  0.517 &\textcolor{gray}{0.653}  &0.000&\cellcolor{cvprblue!15}0.783&  \cellcolor{cvprblue!35}0.907 \\
 & Medium &  0.524 &\textcolor{gray}{0.662}  &0.000&\cellcolor{cvprblue!15}0.866&  \cellcolor{cvprblue!35}0.905   \\
 & Large  &  0.522 &\textcolor{gray}{0.663}  &0.000&\cellcolor{cvprblue!35}0.874&  \cellcolor{cvprblue!15}0.859   \\
 & Average &  0.521 &  \textcolor{gray}{0.659}&0.000&\cellcolor{cvprblue!15}0.841&  \cellcolor{cvprblue!35}0.890 \\
\midrule
\multirow{4}{*}{On-the-Go} 
 & Small  &  0.420 & \textcolor{gray}{0.924} &0.264&\cellcolor{cvprblue!35}0.948&  \cellcolor{cvprblue!15}0.850   \\
 & Medium &  0.429 & \textcolor{gray}{0.923} &0.257&\cellcolor{cvprblue!35}0.940&  \cellcolor{cvprblue!15}0.902   \\
 & Large  &  0.427 & \textcolor{gray}{0.922} &0.261&\cellcolor{cvprblue!35}0.921&  \cellcolor{cvprblue!15}0.899   \\
 & Average &  0.425 &  \textcolor{gray}{0.923}&0.261&\cellcolor{cvprblue!35}0.936&  \cellcolor{cvprblue!15}0.884 \\
\midrule
\multirow{4}{*}{RobustNeRF} 
 & Small  &  0.100 &\textcolor{gray}{0.644}  &0.014&\cellcolor{cvprblue!15}0.546&  \cellcolor{cvprblue!35}0.692   \\
 & Medium &  0.100 &\textcolor{gray}{0.647}  &0.015&\cellcolor{cvprblue!15}0.602&  \cellcolor{cvprblue!35}0.626   \\
 & Large  &  0.112 &\textcolor{gray}{0.650}  &0.014&\cellcolor{cvprblue!35}0.610&  \cellcolor{cvprblue!15}0.606   \\
 & Average &  0.104 &  \textcolor{gray}{0.647}&0.014&\cellcolor{cvprblue!15}0.586&  \cellcolor{cvprblue!35}0.641 \\
\midrule
\multirow{4}{*}{ETH3D} 
 & Small  &  0.292 &  \textcolor{gray}{0.888}&0.029&\cellcolor{cvprblue!35}0.995&  \cellcolor{cvprblue!15}0.885  \\
 & Medium &  0.298 &  \textcolor{gray}{0.891}&0.034&\cellcolor{cvprblue!35}0.987&  \cellcolor{cvprblue!15}0.926  \\
 & Large  &  0.303 &  \textcolor{gray}{0.883}&0.040&\cellcolor{cvprblue!35}0.974&  \cellcolor{cvprblue!15}0.930  \\
 & Average & 0.298 &  \textcolor{gray}{0.887}&0.034&\cellcolor{cvprblue!35}0.985&  \cellcolor{cvprblue!15}0.914 \\
\bottomrule
\end{tabular}}\vspace{-5pt}
\end{table}

While this experiment mirrors common VPR benchmarking, it is noteworthy to mention that the scale of input image quantities is different: VGGT operates on tens to 100 images per batch, whereas VPR systems routinely handle thousands. Nevertheless, the approaches are complementary: VGGT-based view selection can re-rank or filter retrieval candidates, suggesting a practical route toward multi-view–aware place recognition at scale.


\section{Conclusion}
\label{sec:conclusion}
In this work, we investigated robustness in feed-forward 3D reconstruction under noisy, in-the-wild image collections. Our analysis revealed an emergent noise-suppression behavior in modern feed-forward architecture \textit{e.g.,} VGGT. We observe that internal attention and feature representations consistently emphasize spatially and geometrically consistent views while downweighting distractors. Building on this finding, we introduced a training-free view filtering mechanism that ranks and selects context images using a single global threshold. The approach is simple to deploy, complementary to existing pipelines, and generalizes datasets, yielding consistent improvements over retrieval-based baselines while preserving the efficiency advantages of feed-forward reconstruction.

\section*{Acknowledgments}
C.F. was supported in part through NSF grants 2514030 and 2238968 in this work. The visit of J.H. to NYU was through the NYU-KAIST collaboration program.
This research was supported by the ETH AI Center through an ETH AI Center postdoctoral fellowship to Sunghwan Hong.

\clearpage
\newpage

\clearpage
\appendix
\setcounter{page}{1}
\section*{\Large Appendix}

This supplementary material provides additional details and analyses to support the main paper, organized as follows:

\begin{itemize} \item \textbf{Implementation details (\S\ref{sup_sec:impl}):} Specific hyperparameters, experimental settings, and pseudo-code for reproducibility. \item \textbf{Additional Experiments (\S\ref{sup_sec:additional_exp}):} Generalization analysis on the Pi3 architecture and introduction of a variant of our method that combines RobustVGGT-$\mathcal A$ and VGGT-$\mathcal F$. \item \textbf{Qualitative Results (\S\ref{sup_sec:additional_results}):} Visual comparisons of  3D reconstructions and internal attention/feature map visualizations. \item \textbf{Limitations and Future Works (\S\ref{sup_sec:limitation}):} Discussion on current constraints and potential directions for explorations. \end{itemize}

\section{Implementation Details}
\label{sup_sec:impl}

We build our method on top of VGGT. For VGGT-$\mathcal{A}$, we set the attention threshold to $\tau^{\text{attn}} = 0.05$, and for VGGT-$\mathcal{F}$, we set the feature threshold to $\tau^{\text{feat}} = 0.65$. The same thresholds are used for all benchmark datasets. Unless otherwise specified, all experiments are conducted on a single NVIDIA A6000 GPU paired with an AMD EPYC 7543 CPU. PyTorch-style pseudo-code for VGGT-$\mathcal{F}$ and VGGT-$\mathcal{A}$ is provided in Alg.~\ref{alg:VGGTF} and Alg.~\ref{alg:VGGTA}, respectively.

\section{Additional Experiments}
\label{sup_sec:additional_exp}
\begin{figure}[t]
    \centering
    \includegraphics[width=1.0\linewidth]{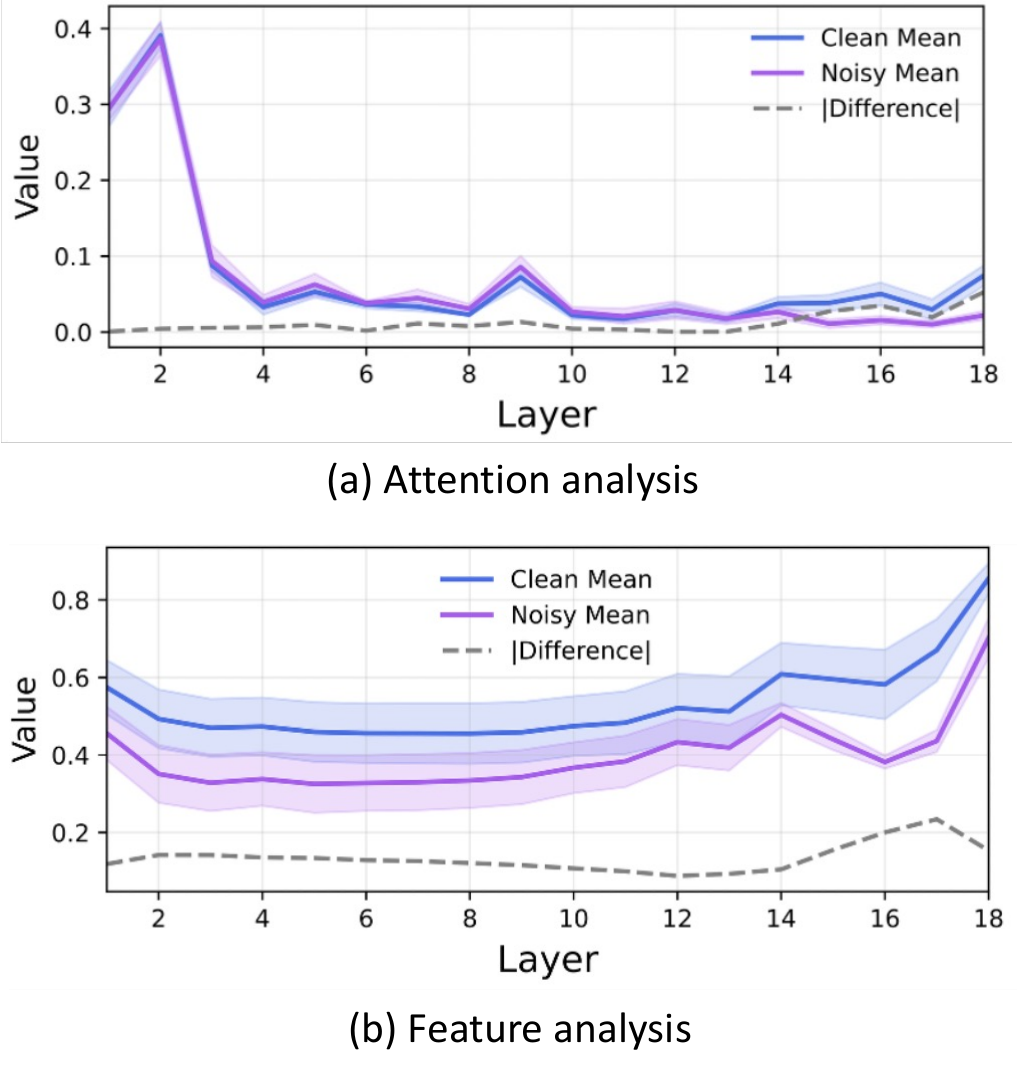}
  \caption{\textbf{Layer-wise analysis of Pi3~\cite{wang2025pi}.} We measure the gap between clean and distractor views for attention and  feature similarity across Pi3’s all layers. The separation grows with depth and peaks at the final layer, indicating emergent noise suppression.}
    \label{fig:pi3_layer}
\end{figure}

\paragraph{Analysis on Pi3.}
In this section, we conduct an additional experiment using Pi3~\cite{wang2025pi}, a preprint that shares a similar architecture as VGGT~\cite{wang2025vggt}. Following the procedure described in our main paper, we first perform a layer-wise analysis, measuring the gap between clean and distractor views in terms of attention and feature similarity. From Fig.~\ref{fig:pi3_layer}, we observe a similar trend to VGGT: the gap steadily increases and peaks in the final layers. Among the layers, layer 17 exhibits the largest difference, and we therefore use this layer for view filtering.

Based on this analysis, we evaluate the effectiveness of the internal representations at this layer. The results are summarized in Tab.~\ref{tab:supp_pose} and Tab.~\ref{tab:supp_depth}. As the noise level increases, our method consistently proves beneficial for both camera pose estimation and multi-view depth estimation.

However, it is noteworthy that as shown in Tab.~\ref{tab:supp_pi3_success_rate}, the success rate of distractor rejection is relatively low. This may explain the smaller gap observed in Fig.~\ref{fig:pi3_layer} compared to the analysis on VGGT.

\paragraph{Blending $\tau^{\mathrm{attn}}$ and $\tau^{\mathrm{feat}}$. }

We further introduce a variant of our method that combines RobustVGGT-$\mathcal{A}$ and RobustVGGT-$\mathcal{F}$, integrating attention-based and feature-based rejection into a single score. This experiment is intended to provide insight into how a simple fusion of the two signals can improve view filtering and motivate more sophisticated variants.

The final score is obtained by first normalizing the attention scores from RobustVGGT-$\mathcal{A}$ and the cosine similarities from RobustVGGT-$\mathcal{F}$, and then aggregating them with a weighted sum:
\begin{equation}
    r_{i\rightarrow j}^{\mathrm{agg}} \;=\; \alpha\, \bar{r}^{\text{att}}_{i\rightarrow j} + (1-\alpha)\, \bar{r}^{\text{feat}}_{i\rightarrow j}, \quad \alpha\in[0,1],
\end{equation}
where $\bar{r}$ denotes the normalized score. The context set for $I_i$ is then defined as
\begin{equation}
    \phi(i) \;=\; \{\, j \;|\; j=i \;\text{ or }\; r_{i\rightarrow j}^{\mathrm{agg}}\ge\tau^{\mathrm{agg}} \,\},
\end{equation}
with a single fixed threshold $\tau^{\mathrm{agg}}$ shared across all benchmarks. We set $\alpha=0.5$ and $\tau^{\mathrm{agg}}=0.4$ in all experiments. As shown in Tab.~\ref{tab:supp_comb_pose}, the combined approach yields more robust performance across all datasets. Furthermore, as can be seen in Tab.~\ref{tab:supp_vggt_success_rate}, the combined method also achieves better performance in terms of success rate.

\section{Additional Qualitative Results}
\label{sup_sec:additional_results}

\begin{figure*}[t]
    \centering
    \includegraphics[width=1.0\linewidth]{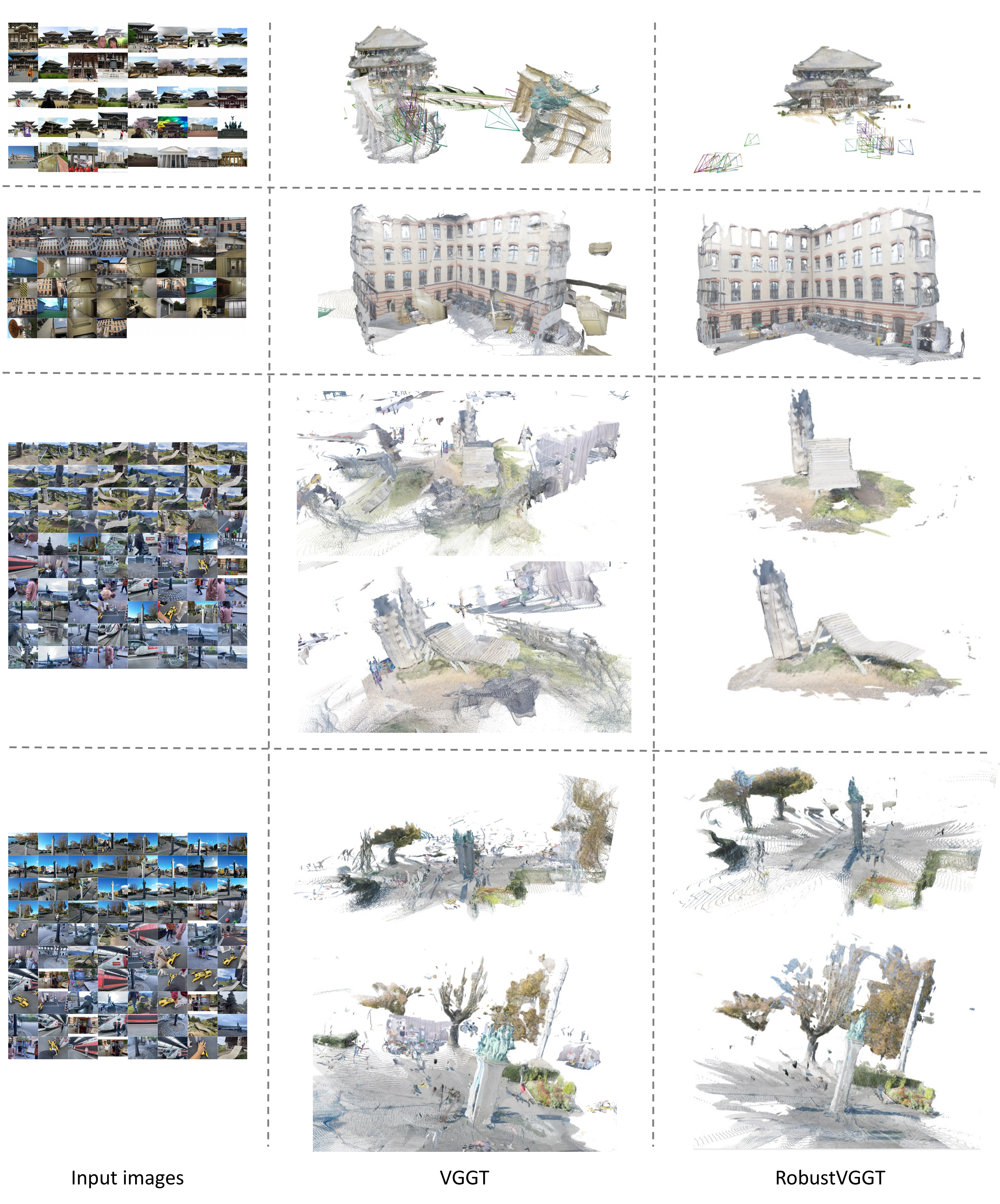}
    \caption{\textbf{Visualization of point maps produced by VGGT and RobustVGGT on various datasets.}} 
    \label{fig:supp_pointmap}
\end{figure*}

\begin{figure*}[t]
    \centering
    \includegraphics[width=1.0\linewidth]{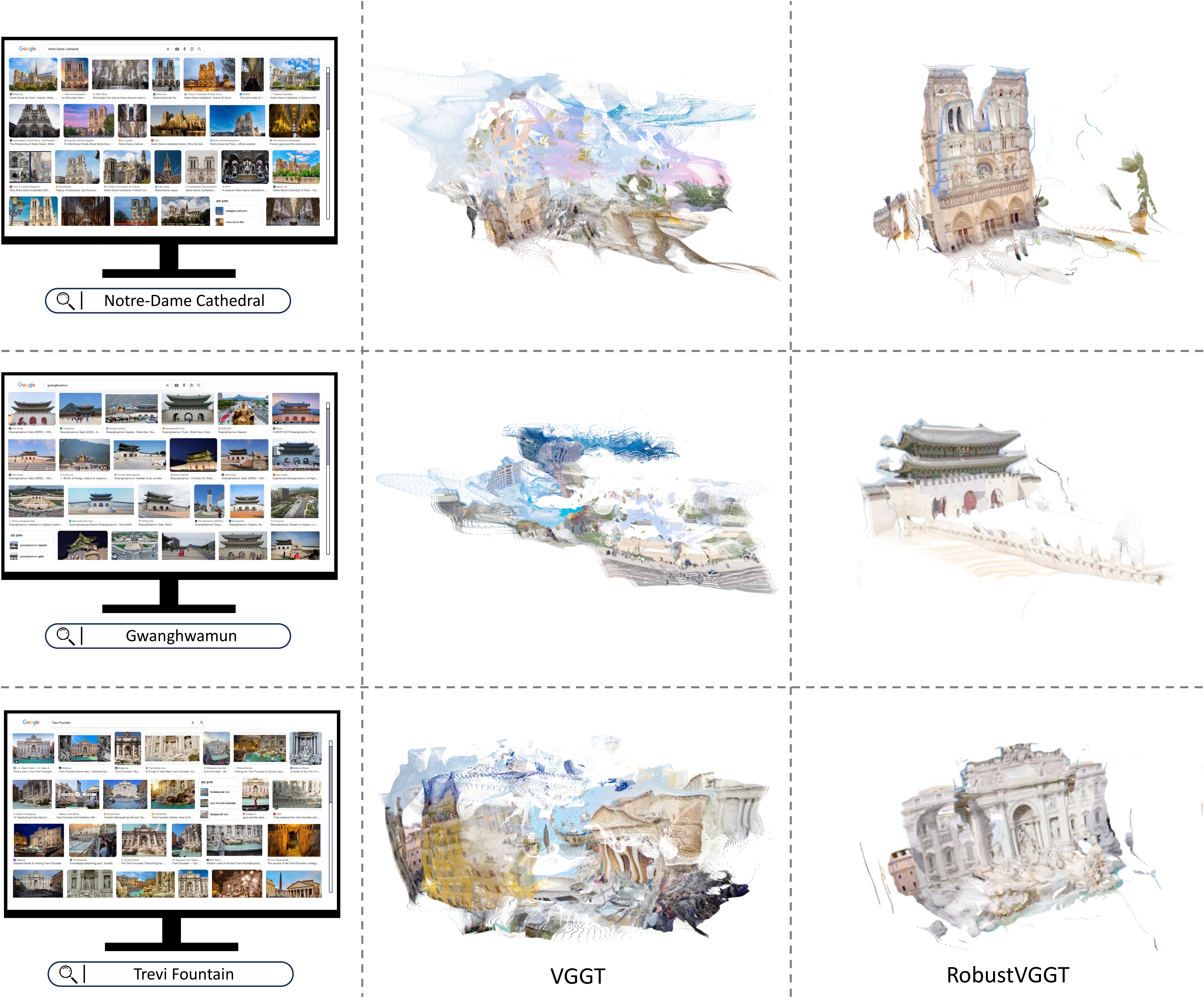}
    \caption{\textbf{Visualization of point maps produced by VGGT and RobustVGGT on internet-collected images.} } 
    \label{fig:supp_internet}
\end{figure*}

We provide more qualitative results in Fig.~\ref{fig:supp_pointmap}, where we show a comparison of reconstructed scenes between our method and VGGT. As can be seen in the reconstructed point maps, VGGT—without filtering out distractor images—yields severely degraded reconstructions even for views corresponding to clean images. Furthermore, as shown in Fig.~\ref{fig:supp_internet}, when reconstructing scenes using landmark images collected from the internet (e.g., Google) in real-world scenarios, VGGT fails to reject distracting or irrelevant images, resulting in inaccurate scene reconstructions. In contrast, our method robustly rejects such images and produces significantly more reliable scene reconstructions.

\section{Additional Visualizations}

\begin{figure*}[t]
    \centering
    \includegraphics[width=0.8\linewidth]{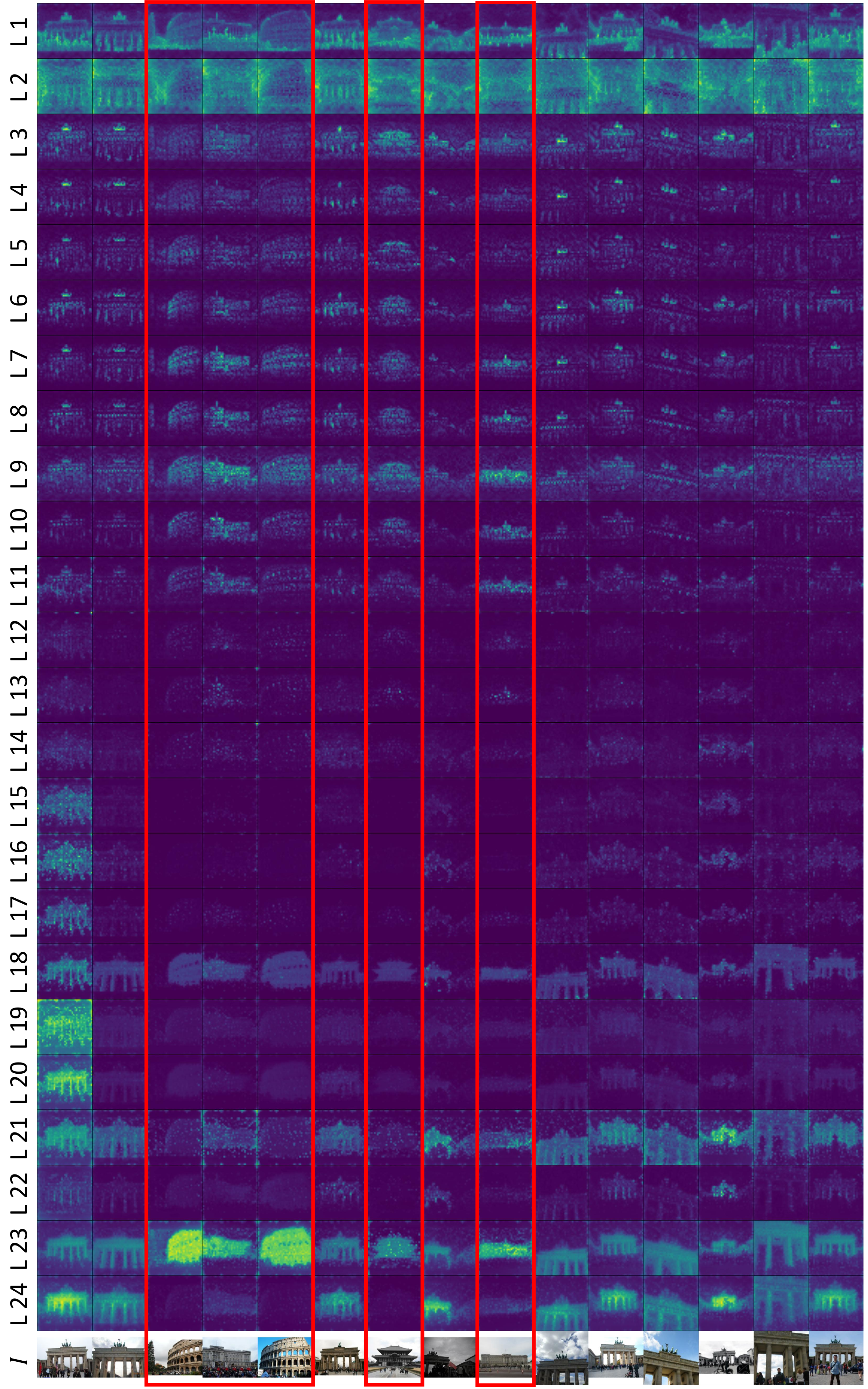}
    \caption{\textbf{Visualization of attention maps across all layers of VGGT on the Phototourism dataset.}} 
    \label{fig:supp_attn1}
\end{figure*}

\begin{figure*}[t]
    \centering
    \includegraphics[width=0.75\linewidth]{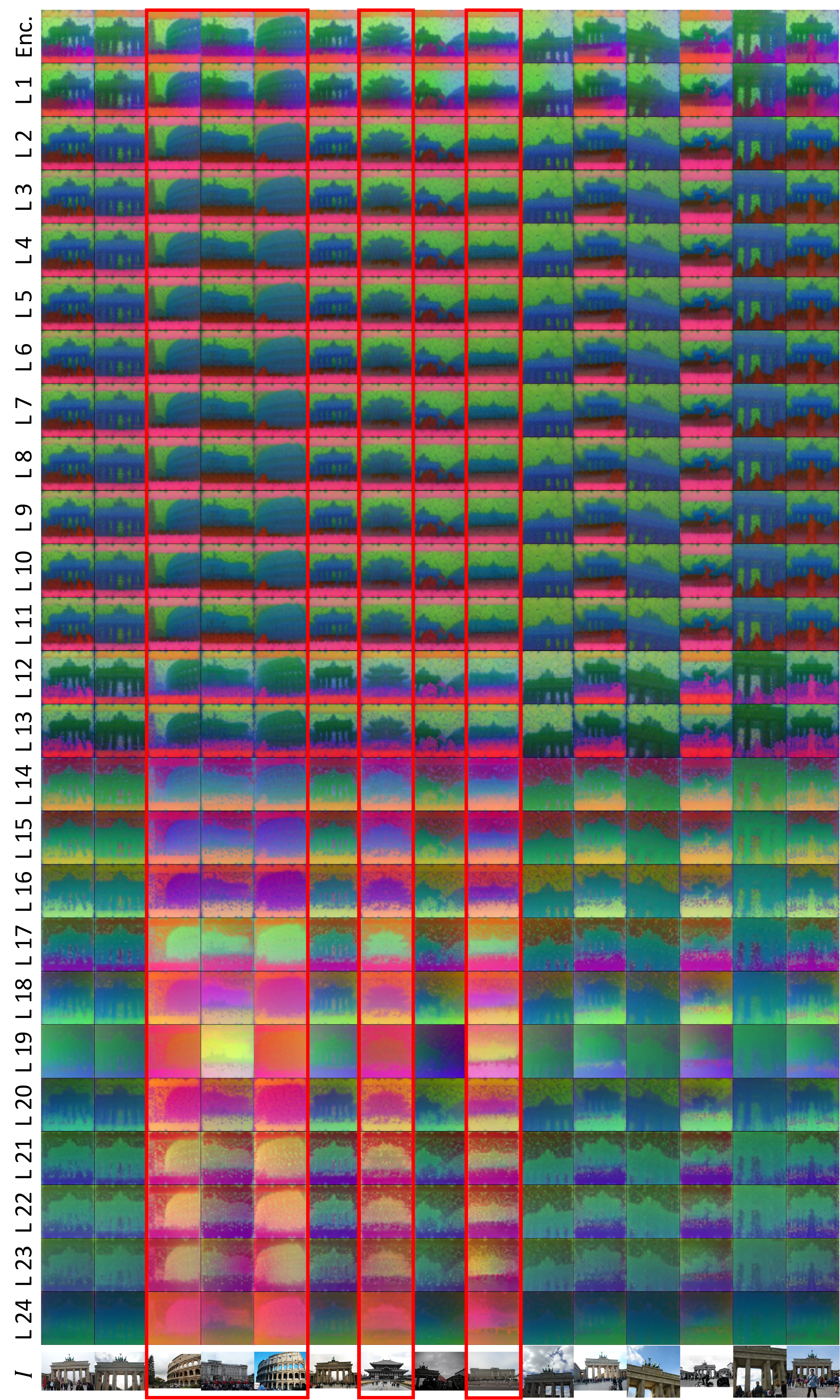}
    \caption{\textbf{Visualization of feature maps of all layers inside VGGT on Phototourism dataset.} \textit{Enc.} denotes features from VGGT's encoder.} 
    \label{fig:supp_feat1}
\end{figure*}

\begin{figure*}[t]
    \centering
    \includegraphics[width=0.75\linewidth]{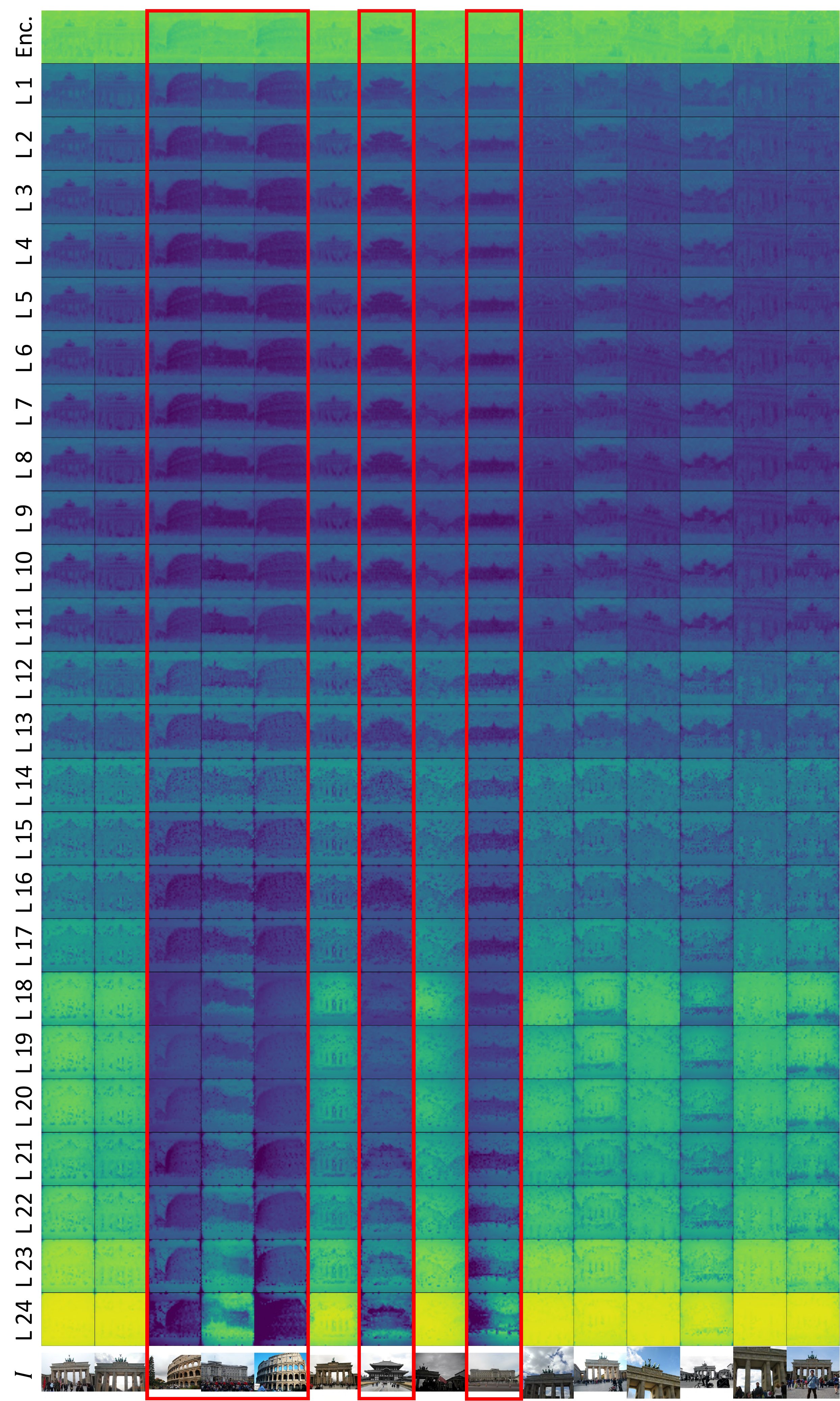}
    \caption{\textbf{Visualization of correlation maps across all layers of VGGT on the Phototourism dataset.} 
\textit{Enc.} denotes features from VGGT's encoder.}

    \label{fig:supp_corr1}
\end{figure*}

\begin{figure*}[t]
    \centering
    \includegraphics[height=1.0\linewidth]{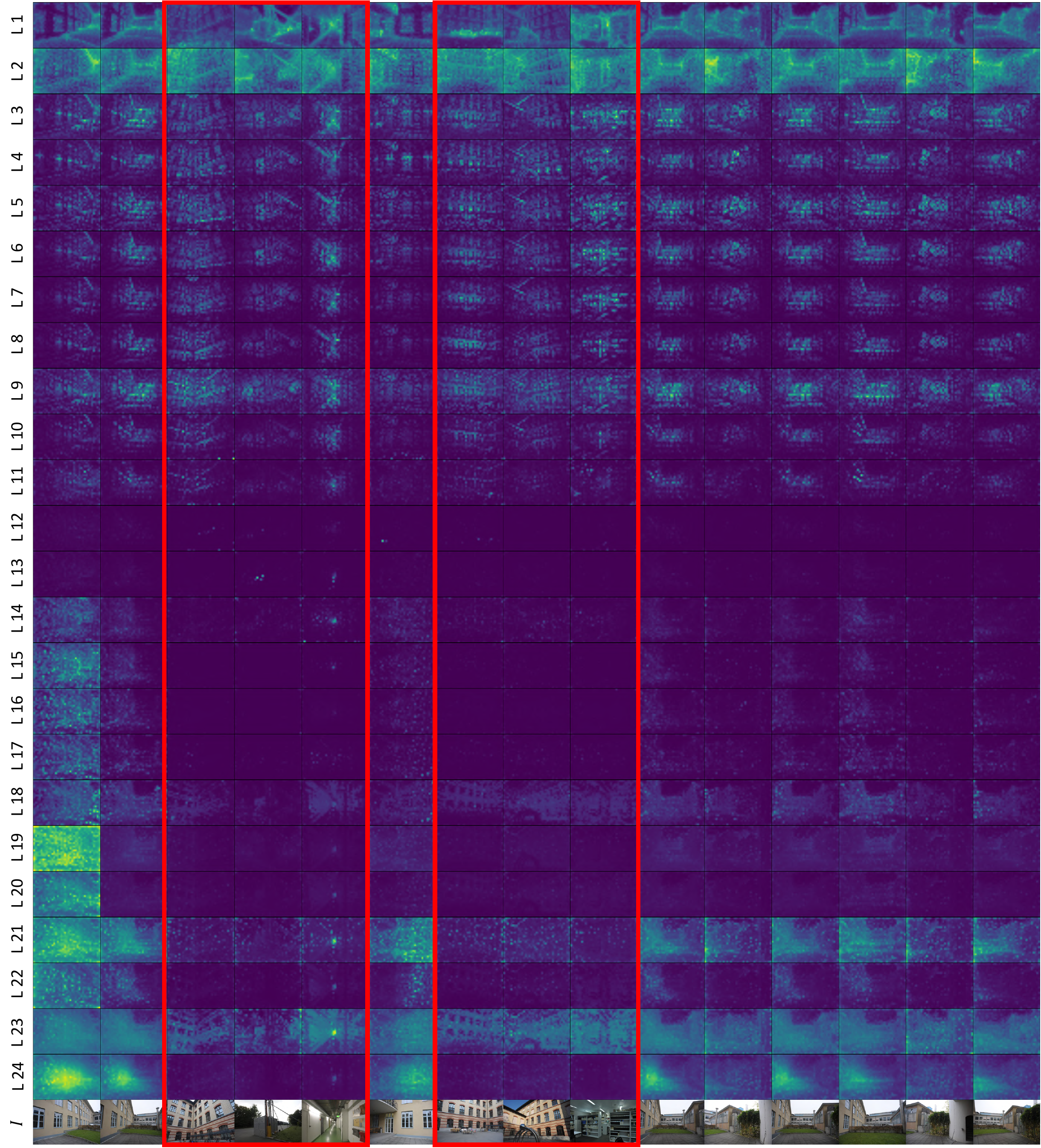}
    \caption{\textbf{Visualization of attention maps across all VGGT layers on the ETH3D dataset.}}

    \label{fig:supp_attn2}
\end{figure*}

\begin{figure*}[t]
    \centering
    \includegraphics[height=1.0\linewidth]{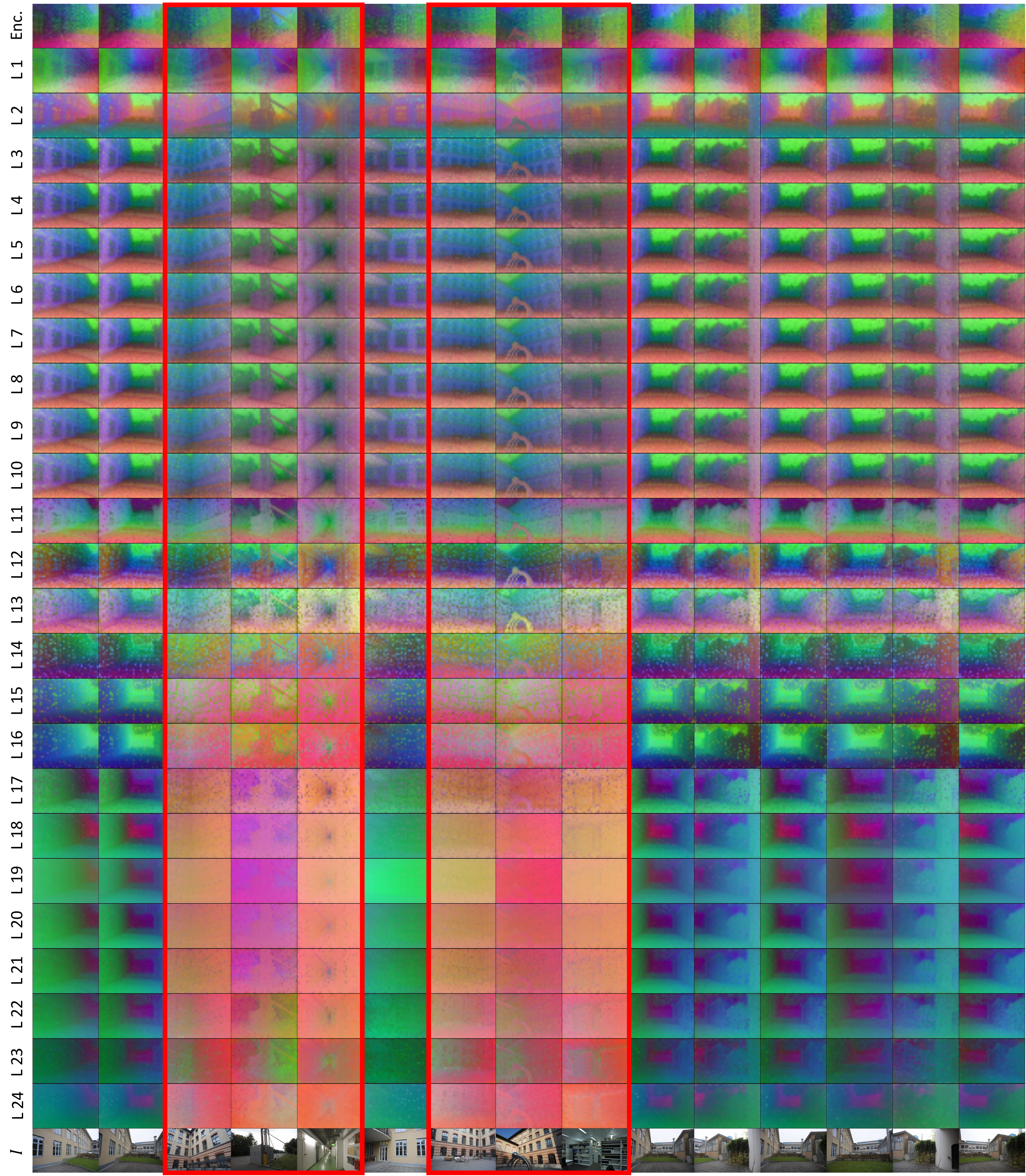}
    \caption{\textbf{Visualization of feature maps across all VGGT layers on the ETH3D dataset.} \textit{Enc.} denotes features from VGGT's encoder.}

    \label{fig:supp_feat2}
\end{figure*}

\begin{figure*}[t]
    \centering
    \includegraphics[height=1.0\linewidth]{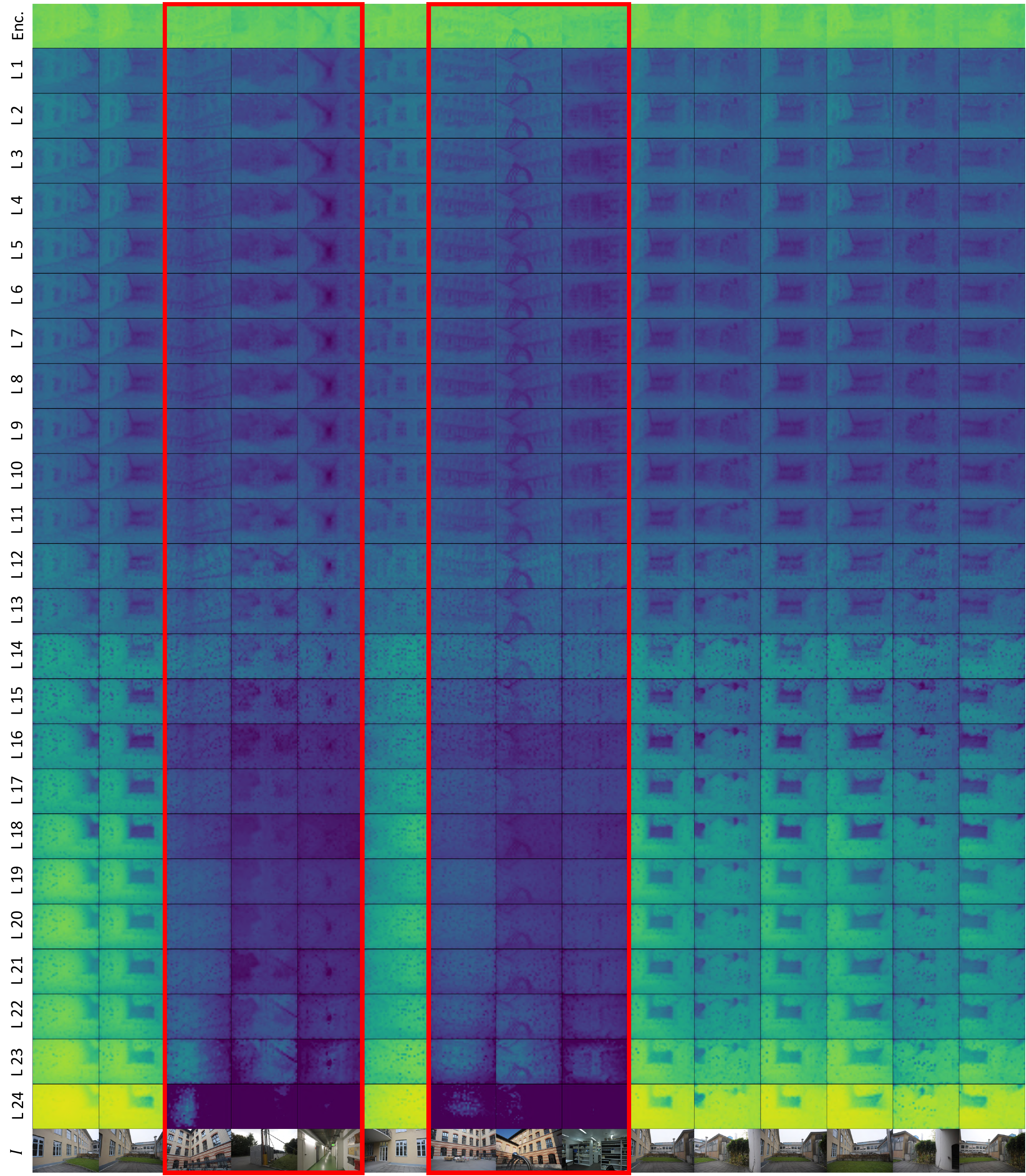}
  \caption{\textbf{Visualization of correlation maps across all VGGT layers on the ETH3D dataset.} \textit{Enc.} denotes features from VGGT's encoder.}

    \label{fig:supp_corr2}
\end{figure*}

We also provide additional visualizations of attention maps, feature maps and correlation maps on Phototurism and ETH3D datasets, in Fig.~\ref{fig:supp_attn1}, Fig.~\ref{fig:supp_feat1}, Fig.~\ref{fig:supp_corr1}, Fig.~\ref{fig:supp_attn2}, Fig.~\ref{fig:supp_feat2} and Fig.~\ref{fig:supp_corr2}. Interestingly, attention visualizations indicate that L23 allocates elevated attention to distractor images. Coupled with the observed suppression at L24, this is consistent with a two-stage behavior in the final layers: L23 exposes or aggregates signals from distractors, and L24 performs the actual rejection. While not proof of a hard specialization, this layer-wise pattern supports our hypothesis of late-layer differentiation between exposure and suppression.

\section{Limitations and Future Works}
\label{sup_sec:limitation}

In this work, we showed that feed-forward 3D geometry foundation models \textit{e.g.,} VGGT, exhibit an \emph{emergent} noise-suppression behavior, which we exploit to build a simple, training-free, architecture-free view-selection module that is complementary to existing pipelines and effective across benchmarks. Our approach has several limitations and offers a few potential explorations: it operates at the \emph{view} level and does not perform token/patch-wise filtering, which could further improve reconstruction but is beyond our current scope and leave it for a future work; it requires two forward passes (scoring then reconstruction), introducing a modest runtime overhead that remains substantially lower than retrieval-style prefiltering that evaluates per image \textit{e.g.,} MegaLoc's \(O(N)\) passes; and its throughput is bounded by the backbone's batching capacity (typically tens to \(\sim\)100 images on an RTX~A6000). These constraints can be mitigated with larger GPUs or lighter backbones, and scaled via hierarchical or cluster-wise filtering (e.g., applying our selector within VPR-formed clusters), offering a practical path to broader deployment without retraining.

\begin{table*}[t]
    \centering
  \caption{\textbf{Camera pose estimation of Pi3}.}
    \label{tab:supp_pose}
    \resizebox{\textwidth}{!}{
    \begin{tabular}{l|ccc|ccc|ccc|ccc}
        \toprule
        \multirow{2}{*}{Methods} &
        \multicolumn{3}{c|}{Small} &
        \multicolumn{3}{c|}{Medium} &
        \multicolumn{3}{c|}{Large} &
        \multicolumn{3}{c}{Avg} \\
        & ATE $\downarrow$& RPE$_\text{trans}$ $\downarrow$& RPE$_\text{rot}$ $\downarrow$& ATE $\downarrow$& RPE$_\text{trans}$ $\downarrow$& RPE$_\text{rot}$ $\downarrow$& ATE $\downarrow$& RPE$_\text{trans}$ $\downarrow$& RPE$_\text{rot}$ $\downarrow$& ATE $\downarrow$& RPE$_\text{trans}$ $\downarrow$& RPE$_\text{rot}$ $\downarrow$\\
        \midrule
        \rowcolor{gray!20} \multicolumn{13}{c}{\textbf{Phototourism}} \\
        \midrule
        Pi3~\cite{wang2025pi}    & 0.2512& 0.3815& 0.9914&  0.3861&  0.6049&  1.8208&  0.5740&  0.8962&  3.1781&  0.4038&  0.6275&  1.9968\\
        \midrule
         RobustPi3-$\mathcal A$ & \cellcolor{cvprblue!15}0.2382& \cellcolor{cvprblue!15}0.3606&\cellcolor{cvprblue!15}0.9211& \cellcolor{cvprblue!35}0.2652&  \cellcolor{cvprblue!35}0.4361& \cellcolor{cvprblue!35}1.2007& \cellcolor{cvprblue!35}0.3174& \cellcolor{cvprblue!35}0.5730& \cellcolor{cvprblue!15}2.6909& \cellcolor{cvprblue!35}0.2736& \cellcolor{cvprblue!35}0.4566& \cellcolor{cvprblue!35}1.6042\\
         RobustPi3-$\mathcal F$& \cellcolor{cvprblue!35}0.2056& \cellcolor{cvprblue!35}0.3084& \cellcolor{cvprblue!35}0.8355& \cellcolor{cvprblue!15}0.3093& \cellcolor{cvprblue!15}0.4749& \cellcolor{cvprblue!15}1.4506& \cellcolor{cvprblue!15}0.4605& \cellcolor{cvprblue!15}0.7110& \cellcolor{cvprblue!35}2.5300& \cellcolor{cvprblue!15}0.3251& \cellcolor{cvprblue!15}0.4981&\cellcolor{cvprblue!15}1.6054\\ 
        \midrule
        \rowcolor{gray!20} \multicolumn{13}{c}{\textbf{On-the-Go}} \\
        \midrule
        Pi3~\cite{wang2025pi}    & 0.0860& 0.1348& 1.1495&  0.1581&  0.2454&  1.8549&  0.4234&  0.6715&  4.5146&  0.2225&  0.3506&  2.5063\\
        \midrule
         RobustPi3-$\mathcal A$ & \cellcolor{cvprblue!15}0.0718& \cellcolor{cvprblue!15}0.1150&\cellcolor{cvprblue!15}0.9588& \cellcolor{cvprblue!15}0.0903& \cellcolor{cvprblue!15}0.1441&  \cellcolor{cvprblue!15}1.0998& \cellcolor{cvprblue!35}0.1506& \cellcolor{cvprblue!15}0.2832& \cellcolor{cvprblue!15}3.1712& \cellcolor{cvprblue!15}0.1042& \cellcolor{cvprblue!15}0.1808& \cellcolor{cvprblue!15}1.7433\\
         RobustPi3-$\mathcal F$& \cellcolor{cvprblue!35}0.0575& \cellcolor{cvprblue!35}0.0947& \cellcolor{cvprblue!35}0.8910& \cellcolor{cvprblue!35}0.0651& \cellcolor{cvprblue!35}0.1050& \cellcolor{cvprblue!35}0.9831& \cellcolor{cvprblue!15}0.1736& \cellcolor{cvprblue!35}0.2701& \cellcolor{cvprblue!35}1.8804& \cellcolor{cvprblue!35}0.0987& \cellcolor{cvprblue!35}0.1566&\cellcolor{cvprblue!35}1.2515\\ 
        \midrule
        \rowcolor{gray!20} \multicolumn{13}{c}{\textbf{RobustNeRF}} \\
        \midrule
        Pi3~\cite{wang2025pi}    & \cellcolor{cvprblue!35}0.1467& \cellcolor{cvprblue!35}0.2983& 1.0091&  0.2433&  0.4555&  3.4803&  0.3204&  0.5974&  4.6081&  0.2368&  0.4504&  3.0325\\
        \midrule
         RobustPi3-$\mathcal A$ & \cellcolor{cvprblue!15}0.1484&\cellcolor{cvprblue!15}0.3024& \cellcolor{cvprblue!15}0.9915&  \cellcolor{cvprblue!15}0.2258&  \cellcolor{cvprblue!15}0.4689&  \cellcolor{cvprblue!15}3.0924&  \cellcolor{cvprblue!15}0.2258& \cellcolor{cvprblue!15}0.4689&  \cellcolor{cvprblue!15}3.0924&  \cellcolor{cvprblue!15}0.2000&  \cellcolor{cvprblue!15}0.4134&  \cellcolor{cvprblue!15}2.3921\\
         RobustPi3-$\mathcal F$& 0.1515& 0.3080& \cellcolor{cvprblue!35}0.9906&\cellcolor{cvprblue!35}0.1576&\cellcolor{cvprblue!35}0.3151& \cellcolor{cvprblue!35}1.0809&\cellcolor{cvprblue!35}0.1813& \cellcolor{cvprblue!35}0.3491& \cellcolor{cvprblue!35}1.4606& \cellcolor{cvprblue!35}0.1635& \cellcolor{cvprblue!35}0.3241&\cellcolor{cvprblue!35}1.1774\\ 
        \midrule
        \rowcolor{gray!20}\multicolumn{13}{c}{\textbf{ETH3D}} \\
        \midrule
        Pi3~\cite{wang2025pi}    & 0.3533& 0.5717& 3.5348&  0.5495&  0.8535&  6.2596&  1.0319&  1.6258&  19.9132&  0.6449&  1.0170&  9.9025\\
        \midrule
         RobustPi3-$\mathcal A$ & \cellcolor{cvprblue!15}0.3275& \cellcolor{cvprblue!15}0.5217& \cellcolor{cvprblue!15}2.5944&  \cellcolor{cvprblue!15}0.4297&  \cellcolor{cvprblue!15}0.6718& \cellcolor{cvprblue!15}3.8187& \cellcolor{cvprblue!15}0.6779&  \cellcolor{cvprblue!15}1.0717& \cellcolor{cvprblue!15}8.0849& \cellcolor{cvprblue!15}0.4784&  \cellcolor{cvprblue!15}0.7551&  \cellcolor{cvprblue!15}4.8327\\
         RobustPi3-$\mathcal F$&\cellcolor{cvprblue!35}0.1954& \cellcolor{cvprblue!35}0.3230& \cellcolor{cvprblue!35}2.5088&\cellcolor{cvprblue!35}0.2906& \cellcolor{cvprblue!35}0.4810& \cellcolor{cvprblue!35}3.7853& \cellcolor{cvprblue!35}0.5922& \cellcolor{cvprblue!35}0.9473& \cellcolor{cvprblue!35}7.6219& \cellcolor{cvprblue!35}0.3594&\cellcolor{cvprblue!35}0.5838& \cellcolor{cvprblue!35}4.6387\\ 
        \bottomrule
    \end{tabular}}
\end{table*}

\begin{table*}[t]
    \centering
  \caption{\textbf{Multi-view depth estimation results of Pi3}.}
    \label{tab:supp_depth}
    \resizebox{0.8\linewidth}{!}{
    \begin{tabular}{l|cccccccc}
        \toprule
        \multirow{2}{*}{Methods} &
        \multicolumn{8}{c}{\cellcolor{gray!20}ETH3D} \\
        & \multicolumn{2}{c}{Small} & \multicolumn{2}{c}{Medium} & \multicolumn{2}{c}{Large} & \multicolumn{2}{c}{Avg}  \\
        & AbsRel $\downarrow$ & $\delta<1.25 \uparrow$ 
        & AbsRel $\downarrow$ & $\delta<1.25 \uparrow$
        & AbsRel $\downarrow$ & $\delta<1.25 \uparrow$
        & AbsRel $\downarrow$ & $\delta<1.25 \uparrow$ \\
        \midrule
        Pi3~\cite{wang2025pi} & 0.0215& 0.9947& 0.0255& 0.9928& 0.0306& 0.9901& 0.0258& 0.9925\\
        \midrule
         RobustPi3-$\mathcal A$ & \cellcolor{cvprblue!15}0.0208& \cellcolor{cvprblue!15}0.9950& \cellcolor{cvprblue!15}0.0233& \cellcolor{cvprblue!15}0.9939& \cellcolor{cvprblue!15}0.0259& \cellcolor{cvprblue!35}0.9929& \cellcolor{cvprblue!15}0.0233&\cellcolor{cvprblue!15}0.9939\\
         RobustPi3-$\mathcal F$ &  \cellcolor{cvprblue!35}0.0188& \cellcolor{cvprblue!35}0.9957&  \cellcolor{cvprblue!35}0.0204& \cellcolor{cvprblue!35}0.9951& \cellcolor{cvprblue!35}0.0254&  \cellcolor{cvprblue!15}0.9928& \cellcolor{cvprblue!35}0.0215&\cellcolor{cvprblue!35}0.9946\\ 
        \bottomrule
    \end{tabular}}
\end{table*}
\begin{table}[t]
\centering
\caption{\textbf{Success rate of distractor rejection of Pi3 across noise levels.}}
\label{tab:supp_pi3_success_rate}
\resizebox{0.87\linewidth}{!}{
\begin{tabular}{
    p{2.5cm}|
    p{2.0cm}|
    >{\centering\arraybackslash}p{2.2cm}
    >{\centering\arraybackslash}p{2.2cm}
}
\toprule
Dataset & Noise level &RobustPi3-$\mathcal F$& RobustPi3-$\mathcal A$ \\
\midrule
\multirow{4}{*}{Phototourism} 
 & Small  &0.605&  0.348 \\
 & Medium &0.408&  0.558 \\
 & Large  &0.285&  0.670 \\
 & Average &0.433&   0.525\\
\midrule
\multirow{4}{*}{On-the-Go} 
 & Small  &0.919&  0.323 \\
 & Medium &0.821&  0.582 \\
 & Large  &0.650&  0.755 \\
 & Average &0.797&   0.553\\
\midrule
\multirow{4}{*}{RobustNeRF} 
 & Small  &0.662&  0.366 \\
 & Medium &0.436&  0.377 \\
 & Large  &0.498&  0.550 \\
 & Average &0.532&   0.431\\
\midrule
\multirow{4}{*}{ETH3D} 
 & Small  &0.938&  0.198 \\
 & Medium &0.754&  0.355 \\
 & Large  &0.496&  0.620 \\
 & Average &0.729&   0.391\\
\bottomrule
\end{tabular}}
\end{table}

\begin{table}[t]
\centering
\caption{\textbf{Success rate of distractor rejection of VGGT across noise levels.}}
\label{tab:supp_vggt_success_rate}
\resizebox{1.0\linewidth}{!}{
\begin{tabular}{l|l|ccc}
\toprule
Dataset & Noise level &RobustVGGT-$\mathcal F$& RobustVGGT-$\mathcal A$ &\makecell{RobustVGGT-$\mathcal A$ \\ + RobustVGGT-$\mathcal F$} \\
\midrule
\multirow{4}{*}{Phototourism} 
 & Small  &0.783&  0.907 &0.968\\
 & Medium &0.866&  0.905   &0.982\\
 & Large  &0.874&  0.859   &0.983\\
 & Average &0.841&   0.890 &0.978\\
\midrule
\multirow{4}{*}{On-the-Go} 
 & Small  &0.948&  0.850   &0.966\\
 & Medium &0.940&  0.902   &0.952\\
 & Large  &0.921&  0.899   &0.943\\
 & Average &0.936&   0.884 &0.954\\
\midrule
\multirow{4}{*}{RobustNeRF} 
 & Small  &0.546&  0.692   &0.734\\
 & Medium &0.602&  0.626   &0.636\\
 & Large  &0.610&  0.606   &0.634\\
 & Average &0.586&   0.641 &0.668\\
\midrule
\multirow{4}{*}{ETH3D} 
 & Small  &0.995&  0.885  &0.997\\
 & Medium &0.987&  0.926  &0.991\\
 & Large  &0.974&  0.930  &0.985\\
 & Average &0.985&   0.914 &0.991\\
\bottomrule
\end{tabular}}
\end{table}

\begin{table*}[t]
    \centering
  \caption{\textbf{Camera pose estimation using the combination of RobustVGGT-$\mathcal{A}$ and RobustVGGT-$\mathcal{F}$.} We report the camera pose evaluation for the RobustVGGT-$\mathcal{A}+F$ extension, which blends $\tau^{\text{attn}}$ from RobustVGGT-$\mathcal{A}$ and $\tau^{\text{feat}}$ from RobustVGGT-$\mathcal{F}$.}
    \label{tab:supp_comb_pose}
    \resizebox{\textwidth}{!}{
    \begin{tabular}{l|ccc|ccc|ccc|ccc}
        \toprule
        \multirow{2}{*}{Methods} &
        \multicolumn{3}{c|}{Small} &
        \multicolumn{3}{c|}{Medium} &
        \multicolumn{3}{c|}{Large} &
        \multicolumn{3}{c}{Avg} \\
        & ATE $\downarrow$& RPE$_\text{trans}$ $\downarrow$& RPE$_\text{rot}$ $\downarrow$& ATE $\downarrow$& RPE$_\text{trans}$ $\downarrow$& RPE$_\text{rot}$ $\downarrow$& ATE $\downarrow$& RPE$_\text{trans}$ $\downarrow$& RPE$_\text{rot}$ $\downarrow$& ATE $\downarrow$& RPE$_\text{trans}$ $\downarrow$& RPE$_\text{rot}$ $\downarrow$\\
        \midrule
        \rowcolor{gray!20}\multicolumn{13}{c}{\textbf{Phototourism}} \\
        \midrule
        VGGT~\cite{wang2025vggt} & 0.3068& 0.4553& 0.9906&  0.3612&  0.5314&  1.1987&  0.3833&  0.5649&  1.3304&  0.3504&  0.5172&  1.1732\\
        \midrule
        RobustVGGT-$\mathcal A$ & 0.2732& 0.4094&0.8719& 0.2792&  0.4153& 0.8806& 0.2930& 0.4349& 0.9310& 0.2818& 0.4199& 0.8945\\
         RobustVGGT-$\mathcal F$& \cellcolor{cvprblue!15}0.2641& \cellcolor{cvprblue!15}0.3936& \cellcolor{cvprblue!15}0.8420& \cellcolor{cvprblue!15}0.2645& \cellcolor{cvprblue!15}0.3949& \cellcolor{cvprblue!15}0.8402& \cellcolor{cvprblue!15}0.2664& \cellcolor{cvprblue!15}0.3973& \cellcolor{cvprblue!15}0.8388& \cellcolor{cvprblue!15}0.2650& \cellcolor{cvprblue!15}0.3953&\cellcolor{cvprblue!15}0.8403\\ 
         RobustVGGT-$\mathcal A + F$ & \cellcolor{cvprblue!35}0.2611&  \cellcolor{cvprblue!35}0.3895& \cellcolor{cvprblue!35}0.8305& \cellcolor{cvprblue!35}0.2610& \cellcolor{cvprblue!35}0.3890&  \cellcolor{cvprblue!35}0.8289& \cellcolor{cvprblue!35}0.2619& \cellcolor{cvprblue!35}0.3904& \cellcolor{cvprblue!35}0.8320& \cellcolor{cvprblue!35}0.2613& \cellcolor{cvprblue!35}0.3896& \cellcolor{cvprblue!35}0.8305\\
        \midrule
        \rowcolor{gray!20} \multicolumn{13}{c}{\textbf{On-the-Go}} \\
        \midrule
        VGGT~\cite{wang2025vggt} & 0.0788& 0.1239& 1.0261&  0.1281&  0.1963&  1.4194&  0.1562&  0.2393&  1.7315&  0.1210&  0.1865&  1.3923\\
        \midrule
        RobustVGGT-$\mathcal A$ & 0.0578& 0.0952&0.8800& 0.0790& 0.1253&  1.2922& 0.0697& 0.1126& 1.0361& 0.0688& 0.1110& 1.0694\\
         RobustVGGT-$\mathcal F$& \cellcolor{cvprblue!15}0.0521& \cellcolor{cvprblue!15}0.0861& \cellcolor{cvprblue!15}0.8179& \cellcolor{cvprblue!15}0.0568& \cellcolor{cvprblue!15}0.0931& \cellcolor{cvprblue!15}0.8914&\cellcolor{cvprblue!15}0.0660& \cellcolor{cvprblue!15}0.1055& \cellcolor{cvprblue!15}0.9872& \cellcolor{cvprblue!15}0.0583& \cellcolor{cvprblue!15}0.0949&\cellcolor{cvprblue!15}0.8988\\ 
         RobustVGGT-$\mathcal A + F$ & \cellcolor{cvprblue!35}0.0519& \cellcolor{cvprblue!35}0.0858&  \cellcolor{cvprblue!35}0.8146& \cellcolor{cvprblue!35}0.0566&  \cellcolor{cvprblue!35}0.0924&  \cellcolor{cvprblue!35}0.8793& \cellcolor{cvprblue!35}0.0614& \cellcolor{cvprblue!35}0.0993& \cellcolor{cvprblue!35}0.9550& \cellcolor{cvprblue!35}0.0566& \cellcolor{cvprblue!35}0.0925& \cellcolor{cvprblue!35}0.8830\\
        \midrule
        \rowcolor{gray!20}\multicolumn{13}{c}{\textbf{RobustNeRF}} \\
        \midrule
        VGGT~\cite{wang2025vggt} & 0.1519& 0.2742& 1.2052&  0.1598&  0.2908&  1.2311&  0.1680&  0.3062&  1.3493&  0.1599&  0.2904&  1.2619\\
        \midrule
        RobustVGGT-$\mathcal A$ & \cellcolor{cvprblue!35}0.1361&\cellcolor{cvprblue!35}0.2433& \cellcolor{cvprblue!15}1.1766&  0.1379&  0.2447&  \cellcolor{cvprblue!35}1.1657&  0.1406& 0.2478&  1.1632&  0.1382&  0.2453&  1.1685\\
         RobustVGGT-$\mathcal F$& 0.1388& 0.2480& \cellcolor{cvprblue!35}1.1656&\cellcolor{cvprblue!15}0.1374&\cellcolor{cvprblue!15}0.2432& 1.1670&\cellcolor{cvprblue!15}0.1374& \cellcolor{cvprblue!15}0.2415& \cellcolor{cvprblue!35}1.1514& \cellcolor{cvprblue!15}0.1379& \cellcolor{cvprblue!15}0.2442&\cellcolor{cvprblue!35}1.1613\\ 
         RobustVGGT-$\mathcal A + F$ & \cellcolor{cvprblue!35}0.1361& \cellcolor{cvprblue!15}0.2435& 1.1783& \cellcolor{cvprblue!35}0.1363& \cellcolor{cvprblue!35}0.2413& \cellcolor{cvprblue!15}1.1664& \cellcolor{cvprblue!35}0.1365& \cellcolor{cvprblue!35}0.2397&  \cellcolor{cvprblue!15}1.1577& \cellcolor{cvprblue!35}0.1363& \cellcolor{cvprblue!35}0.2415& \cellcolor{cvprblue!15}1.1675\\
        \midrule
        \rowcolor{gray!20} \multicolumn{13}{c}{\textbf{ETH3D}} \\
        \midrule
        VGGT~\cite{wang2025vggt} & 0.8572& 1.3675& 6.3908&  0.9182&  1.5028&   9.6272&  1.0165&  1.7348&  15.2774&  0.9306&  1.5350&  10.4318\\
        \midrule
        RobustVGGT-$\mathcal A$ & 0.7447& 1.1724& 3.8938&  \cellcolor{cvprblue!35}0.7673&  \cellcolor{cvprblue!35}1.2123& 3.7325& \cellcolor{cvprblue!35}0.6874&  \cellcolor{cvprblue!35}1.0708& 4.4779& \cellcolor{cvprblue!35}0.7331&  \cellcolor{cvprblue!35}1.1518&  4.0347\\
         RobustVGGT-$\mathcal F$&\cellcolor{cvprblue!35}0.6224& \cellcolor{cvprblue!35}1.0300& \cellcolor{cvprblue!35}2.7304&0.8038& 1.3159& \cellcolor{cvprblue!35}2.9959& 0.8636& 1.3882& \cellcolor{cvprblue!15}3.4866& 0.7633&1.2447 & \cellcolor{cvprblue!15}3.0710\\ 
         RobustVGGT-$\mathcal A + F$ & \cellcolor{cvprblue!15}0.6521& \cellcolor{cvprblue!15}1.0729& \cellcolor{cvprblue!15}2.7305& \cellcolor{cvprblue!15}0.7847& \cellcolor{cvprblue!15}1.2671& \cellcolor{cvprblue!15}3.1248& \cellcolor{cvprblue!15}0.7861&  \cellcolor{cvprblue!15}1.2534& \cellcolor{cvprblue!35}3.2919& \cellcolor{cvprblue!15}0.7410& \cellcolor{cvprblue!15}1.1978& \cellcolor{cvprblue!35}3.0491\\
        \bottomrule
    \end{tabular}}
\end{table*}

\clearpage
\begin{algorithm*}[t]
\caption{Pseudo-Code of RobustVGGT-$\mathcal F$, PyTorch-like}
\label{alg:VGGTF}
\begin{lstlisting}[style=vscode]
class VGGT_F:
    def forward(images, image_hw):
        predictions, aggregated_tokens_list = VGGT(images)

        target_layer = 23         
        aggregated_tokens_selected = aggregated_tokens_list[target_layer]
        feature = aggregated_tokens_selected[..., 1024:]

        H, W = image_hw
        h_patches = H // patch_size
        w_patches = W // patch_size
        num_patch_tokens = h_patches * w_patches

        feature = feature[:, :, patch_start_idx:, :]   # (B, N, T, C)
        B, N, T, C = feature.shape
        layer_feat = feature.reshape(B * N, T, C)

        ref_feat       = layer_feat[0:1]               
        ref_feat_norm  = normalize(ref_feat)
        layer_feat_norm = normalize(layer_feat)

        cos_sim = einsum(layer_feat_norm, ref_feat_norm)   # (B*N, T, T)
        cos_sim_mean = cos_sim.mean(-1).mean(-1)            # (B*N,)

        filtered_images = filter(images, cos_sim_mean, threshold=0.65)
        predictions, _ = VGGT(filtered_images)

    return predictions
\end{lstlisting}
\end{algorithm*}
\clearpage

\begin{algorithm*}[t]
\caption{Pseudo-Code of RobustVGGT-$\mathcal A$, PyTorch-like}
\label{alg:VGGTA}
\begin{lstlisting}[style=vscode]
class VGGT_A:
    def forward(images, image_hw):
        predictions, Q_layers, K_layers = VGGT(images)

        H, W = image_hw
        h_patches = H // patch_size
        w_patches = W // patch_size
        num_patch_tokens = h_patches * w_patches

        Q = Q_layers[23]                      # (B, H, T, D)
        K = K_layers[23]                      # (B, H, T, D)
        T = int(K.shape[-2])
        num_images_in_seq = T // tokens_per_image

        q_first = Q[:, :, first_image_patch_start:first_image_patch_end, :]
        Tk = int(num_images_in_seq * tokens_per_image)
        K_slice = K[:, :, :Tk, :]
        scale  = 1.0 / sqrt(dim(q_first))
        logits = einsum(q_first, K_slice) * scale      # (B, H, Nq, Tk)
        probs  = softmax(logits, dim=-1)
        attn_first = probs.mean(dim=1).mean(dim=1)[0]  # (Tk,)

        maps_2d = []
        for img_idx in range(num_images_in_seq):
            start = img_idx * tokens_per_image + patch_start_idx
            end   = start + num_patch_tokens
            if start >= len(attn_first):
                break

            patch_attn = attn_first[start:end]
            attn2d = patch_attn.view(h_patches, w_patches)
            maps_2d.append(attn2d)

        global_norm = global_norm(maps2d)
        global_mean_val = float(global_norm.mean())

        filtered_images = filter(images, global_mean_val, threshold=0.05)
        predictions, _, _ = VGGT(filtered_images)

        return predictions
\end{lstlisting}
\end{algorithm*}


\clearpage
{
    \small
    \bibliographystyle{ieeenat_fullname}
    \bibliography{main}
}


\end{document}